\def\year{2020}\relax
\documentclass[letterpaper]{article} % DO NOT CHANGE THIS
\usepackage{aaai20}  % DO NOT CHANGE THIS
\usepackage{times}  % DO NOT CHANGE THIS
\usepackage{helvet} % DO NOT CHANGE THIS
\usepackage{courier}  % DO NOT CHANGE THIS
\usepackage[hyphens]{url}  % DO NOT CHANGE THIS
\usepackage{graphicx} % DO NOT CHANGE THIS
\graphicspath{ {figures/} }
\usepackage{graphicx}  % DO NOT CHANGE THIS
\usepackage{enumitem}
\usepackage{amsmath}
\usepackage{amsfonts}
\usepackage{mathrsfs}
\usepackage{algpseudocode, algorithm}
\usepackage{booktabs}
\usepackage{subcaption}
\usepackage{pdfrender}
\newcommand*{\boldgreek}[1]{%
	\textpdfrender{%
		TextRenderingMode=FillStroke,%
		LineWidth=.35pt,%
	}{#1}%
} 

\DeclareMathOperator*{\argmin}{argmin}
\usepackage{theorem}
\theorembodyfont{\rm}
\newtheorem{proposition}{Proposition}
\usepackage{pdfpages}

\title{Graph-Based Decoding Model for Functional Alignment of Unaligned fMRI Data}
\author{Weida Li,\textsuperscript{\rm 1}
	Mingxia Liu,\textsuperscript{\rm 1,2*}
	Fang Chen,\textsuperscript{\rm 1}
	Daoqiang Zhang\textsuperscript{\rm 1*}\\
\textsuperscript{\rm 1}College of Computer Science and Technology \& MIIT Key Laboratory of Pattern Analysis and Machine Intelligence,\\ Nanjing University of Aeronautics and Astronautics, Nanjing, China\\
\textsuperscript{\rm 2}School of Information Science and Technology, Taishan University, Taian, China\\
\textsuperscript{\rm *}Corresponding Authors:  \{mingxialiu, dqzhang\}@nuaa.edu.cn 
}

\begin{document}

\maketitle

\begin{abstract}
	Aggregating multi-subject functional magnetic resonance imaging (fMRI) data is indispensable for generating valid and general inferences from patterns distributed across human brains. The disparities in anatomical structures and functional topographies of human brains warrant aligning fMRI data across subjects. However, the existing functional alignment methods cannot handle well various kinds of fMRI datasets today, especially when they are not temporally-aligned, i.e., some of the subjects probably lack the responses to some stimuli, or different subjects might follow different sequences of stimuli. In this paper, a cross-subject graph that depicts the (dis)similarities between samples across subjects is used as a priori for developing a more flexible framework that suits an assortment of fMRI datasets. However, the high dimension of fMRI data and the use of multiple subjects makes the crude framework time-consuming or unpractical. To address this issue, we further regularize the framework, so that a novel feasible kernel-based optimization, which permits non-linear feature extraction, could be theoretically developed. Specifically, a low-dimension assumption is imposed on each new feature space to avoid overfitting caused by the high-spatial-low-temporal resolution of fMRI data. Experimental results on five datasets suggest that the proposed method is \emph{not only} superior to several state-of-the-art methods on temporally-aligned fMRI data, \emph{but also} suitable for dealing with temporally-unaligned fMRI data.
\end{abstract}

\section{Introduction}

Functional Magnetic Resonance Imaging (fMRI) is an imaging technology used to measure neural activity by using the blood-oxygen-level-dependent (BOLD) contrast as an indicator for cognitive states \cite{logothetis2002neural}. The informative patterns encoded in fMRI enable investigators to study how the human brain works \cite{haxby2014decoding}. Specifically, the use of multi-subject fMRI data is indispensable for accessing the validity and generality of the findings across subjects \cite{talairach1988co,watson1993area}. From another angle, aggregating multi-subject fMRI data is also critical due to the high-spatial-low-temporal (HSLT) resolution of fMRI, i.e., the number of samples (time points or volumes) is generally much smaller than the number of features (voxels) per subject. However, such aggregation is facing a challenge that both anatomical structure and functional topography vary across subjects \cite{haxby2011common}. Hence, inter-subject alignment is an indispensable step in fMRI analysis.

Existing studies for inter-subject alignment include anatomical alignment and functional alignment, which can work in unison. In fact, anatomical alignment is usually used as a preprocessing step for fMRI analysis, by aligning anatomical features based on structural MRI images across subjects. Typical examples include Talairach alignment \cite{talairach1988co}, cortical surface alignment \cite{fischl1999high} and so on. However, anatomical alignment generated limited accuracy since the size, shape and anatomical location of functional loci differ across subjects \cite{watson1993area,rademacher1993topographical}. In contrast, functional alignment tries to directly align functional responses across subjects \cite{sabuncu2009function,conroy2009fmri}. As more radical approaches of functional alignment, Hyperalignment \cite{haxby2011common} and Shared Response Model (SRM) \cite{chen2015reduced} learn implicitly shared patterns across subjects, which are closely related to multi-view Canonical Correlation Analysis (CCA). Though both of them have been extensively studied and extended to an assortment, the existing related studies assume that the given fMRI datasets should be temporally-aligned across subjects \cite{chen2015reduced,turek2018capturing,xu2012regularized}. In other words,  the sequential fMRI time points of each subject have to correspond to the same sequence of stimuli, like all subjects watching a movie together. Such a demand makes them not flexible enough as fMRI datasets today could be not temporally-aligned. For example, some subjects probably lack the responses to some stimuli, or different subjects may respond to different sequences of stimuli. Even though this problem could be somewhat solved by reordering and truncating (or down-sampling) the dataset to generate an aligned version \cite{chen2015reduced}, these processes may lead to an inevitable loss of information. A recent study tries to extend SRM into a semi-supervised one by exploiting labeled samples, the unlabeled samples are, however, required to be temporally-aligned~\cite{turek2017semi}.

In this paper, we aim to develop an adaptable functional alignment framework by using a cross-subject graph that depicts the (dis)similarities between all samples as a priori. Such a graph can be generated according to samples' category labels or through inference~\cite{de2010multi}. From this perspective, we can then focus on the (dis)similarity, which is encoded in a graph, between any two samples rather than merely caring about if the given fMRI dataset is temporally-aligned. However, the crude framework is unpractical as the related matrices are too large to be used, which is caused by the high dimension of fMRI data and the use of multiple subjects. To address this problem, the unrefined framework is regularized so that a novel feasible kernel-based optimization, which allows for non-linear feature extraction, could be theoretically set up. With such a regularization, the optimal solution is, sometimes, unique. Nevertheless, the high-spatial-low-temporal (HSLT) resolution of fMRI data causes that the generated optimal solution could indicate overfitting, i.e., it aligns all aligning samples perfectly. In a specific case, the culprit is that the dimension of the subspace spanned by the aligning samples equals to the number of them. Therefore, a low-dimension assumption, which agrees with that the number of informative features is generally less than the number of voxels, is imposed on each new feature space to avoid overfitting. The refined framework, together with the proposed optimization method, is referred to as Graph-based Decoding Model (GDM) in this paper. Notably, the objective function of Hyperalignment is the same as that of our GDM (with an evident graph). The main contributions of this paper are summarized as follows:
\begin{enumerate}[label=\roman*)]
	
	\item Unlike previous studies that rely on temporally-aligned data, GDM does not require temporal alignment for fMRI data. Once the prior information of the (dis)similarities among samples is available or can be inferred, one can employ our GDM to solve fMRI-based problems at hand. %comes into play.
	
	\item Different from the conventional naive kernel implementation, the computational time of our proposed kernel-based optimization (naturally accompanied by a low-dimension assumption) is faster on the number of samples, making it suitable for processing large-scale dataset.
	
	\item The feasible kernel-based optimization method with the low-dimension assumption is equipped with some theoretical guarantees.   
\end{enumerate}

In the following, we first briefly review related works, and concisely mention the notation and problem statement. Then, the proposed method GDM will be introduced in detail. We further introduce the materials used in this work, experimental setup, competing methods, and experimental results achieved by different methods on both aligned and unaligned datasets, which is followed by a Conclusion section. Related proofs and additional experimental results are given in the \emph{Supplementary File}.

\section{Related Works}
The initial Hyperalignment (HA) method aims to seek implicitly shared features across subjects \cite{haxby2011common}, which is based on the orthogonal Procrustes problem. It is the first that links functional alignment and multi-view CCA. The performance of Hyperalignment on fMRI analysis is dramatically increased compared with any other anatomical alignment methods. To tackle the singularity caused by the HSLT resolution of fMRI, Regularized Hyperalignment (RHA) was developed by Xu \textit{et al.}\ \cite{xu2012regularized}.

However, neither of HA nor RHA can handle full-brain data. To address this issue, there have been several works: Chen \textit{et al.}\ developed a Singular Vector Decomposition Hyperalignment (SVDHA), which firstly carries out a joint-SVD by grouping all subjects' fMRI data for dimension reduction across subjects \cite{chen2014joint}. Later, Chen \textit{et al.}\ introduced a Shared Response Model (SRM) which can be modeled from probabilistic perspective by assuming that each sample from the latent common space has undergone a Gaussian noise disturbance \cite{chen2015reduced}. Solely linear feature extraction was considered until that Kernel Hyperalignment (KHA) was formulated by Lorbert and Ramadge \cite{lorbert2012kernel}. Since fMRI dataset may partially contain labels, a semi-supervised scheme based on SRM was studied by Turek \textit{et al.}\ \cite{turek2017semi}.

On the other hand, a Searchlight approach, which takes functional alignment method as a module, was established to enhance functional alignment further by assuming that any voxel is only in connection with voxels in its anatomical vicinity \cite{guntupalli2016model}. Recently, a Robust SRM that accounts for individual variations was developed by Turek \textit{et al.}\ \cite{turek2018capturing}.

\section{Notation and Problem Statements}
\paragraph{Notation}
In this paper, the bold letters are reserved for matrices (upper) or vectors (lower), whereas the plain are for scalars. Given any sequence of matrices $ \{ \mathbf{A}_{i} \}_{i=1}^{M} $, let $ \mathbf{A}_{*} $ be the corresponding block diagonal matrix whose diagonal matrices are $ \{ \mathbf{A}_{i} \}_{i=1}^{M} $ from the top left to the bottom right. Plus, for any matrix $ \mathbf{A} $, $ \mathbf{a}_{i} $ refers to its $ i $-th column vector, $ A_{ij} $ is its $ (i,j)$-th entry, $ \mathrm{R}(\mathbf{A}) $ denotes the subspace spanned by the columns of $ \mathbf{A} $ and $ \mathrm{N}(\mathbf{A}) $ is the null space of $ \mathbf{A} $, i.e., $ \{ \mathbf{x}\,|\,\mathbf{A}\mathbf{x}=\mathbf{0} \} $. Moreover, any vector is treated as a column vector and the subscript of $ \mathbf{A}_{I\times J} $ indicates its shape.

Let $ \{ \mathbf{X}_{i} \in \mathbb{R}^{V_{i} \times T_{i}} \}_{i=1}^{M} $ be an fMRI dataset where $ T_{i} $ and $ V_{i} $ are the number of samples (time points or volumes) and features (voxels) of the $ i $-th subject, respectively, and $ M $ is the total number of subjects. Due to the HSLT resolution of fMRI, $ T_{i} \ll V_{i} $. To develop a kernel-based method, we introduce a column-wised non-linear map $ \Phi_{i} $ that maps each sample, e.g., each column of $ \mathbf{X}_{i} $, of the $ i $-th subject into a new feature space $ \mathcal{H}_{i} $, which is a Hilbert space. Unlike Kernel Hyperalignment \cite{lorbert2012kernel}, subject-specific kernels are allowed. Here, different kernels could be thought to account for different structures of human brain. For simplicity, denote $ \mathbf{\Phi}_{i} $ by setting $ (\boldgreek{\phi}_{i})_{j} = \Phi_{i}((\mathbf{x}_{i})_{j}) $ for $ 1\leq j\leq T_{i} $, and let $ \mathbf{K}_{i} $ be $ \mathbf{\Phi}_{i}^{T}\mathbf{\Phi}_{i} $. Plus, let $ K $ denote the number of the shared features across subjects.

\paragraph{Assumption for Theoretical Development}
Generally, the dimension of $ \mathcal{H}_{i} $ could be infinite. For example, the reproducing kernel Hilbert space of Gaussian kernel is isomorphic to a subspace of $ l_{2}(\mathbb{N}) $ \cite{steinwart2008support}. For clarity in the development of the optimization, we assume that $ \mathcal{H}_{i} $ is a finite dimensional real Hilbert space throughout the paper. The general lengthy proofs are left in the \emph{Supplementary File}. Thus, $ \Phi_{i}:\mathbb{R}^{V_{i}} \mapsto \mathbb{R}^{N_{i}} $ and $ \mathbf{\Phi}_{i} \in \mathbb{R}^{N_{i} \times T_{i}} $ where $ N_{i} $ is the dimension of $ \mathcal{H}_{i} $.

The goal is to learn aligning maps $ \{ f_{i}: \mathbb{R}^{V_{i}} \mapsto \mathbb{R}^{K} \}_{i=1}^{M} $ for each subject such that they map populations of subjects' fMRI responses into a shared space in which the disparities between subjects' brains are eliminated. Here, we aim to learn linear aligning maps $ \{ h_{i}: \mathbb{R}^{N_{i}} \mapsto \mathbb{R}^{K} \} $ with good generalization. Therefore, $ f_{i} = h_{i}\circ \Phi_{i} $ and $ h_{i}((\boldgreek{\phi}_{i})_{j}) = \mathbf{W}_{i}^{T}(\boldgreek{\phi}_{i})_{j} $ for $ 1\leq j\leq T_{i} $ where $ \mathbf{W}_{i} \in \mathbb{R}^{N_{i}\times K} $.

\section{Proposed Method}
\subsection{Formulation} 
\paragraph{Cross-Subject Graph} A graph about the (dis)similarities among all samples are mostly available. For example, the part of temporally-aligned samples, the category of each sample, or the distances between samples tell which samples are closely related or distinctive. To describe such (dis)similarities, let $ \mathbf{G} \in \mathbb{R}^{T\times T} $ be a cross-subject graph matrix where $ T = \sum_{i=1}^{M}T_{i} $ and $ G_{ij} $ indicates the (dis)similarity of the $ i $-th and $ j $-th samples, and thus $ \mathbf{G}^{T} = \mathbf{G} $. Here, $ i $ or $ j $ could refer to any sample from any subject.

\paragraph{Objective Function}
Let $ \mathbf{W}^{T} $ be $ \begin{pmatrix}
\mathbf{W}_{1}^{T} & \cdots & \mathbf{W}_{M}^{T}
\end{pmatrix} $ and $ \mathbf{Y} $ be $ \mathbf{W}^{T}\mathbf{\Phi}_{*} = \begin{pmatrix}
\mathbf{W}_{1}^{T}\mathbf{\Phi}_{1} & \cdots & \mathbf{W}_{M}^{T}\mathbf{\Phi}_{M} 
\end{pmatrix} $. Since $ \mathbf{Y} \in \mathbb{R}^{K\times T} $ contains all samples, the objective function can be expressed as
\begin{gather}
\argmin_{\mathbf{W}}\cfrac{1}{2}\sum_{i=1}^{T}\sum_{j=1}^{T}G_{ij}\left\lVert \mathbf{y}_{i}-\mathbf{y}_{j} \right\rVert_{F}^{2}  = \mathrm{tr}\left( \mathbf{Y}\mathbf{L}\mathbf{Y}^{T} \right)
\end{gather}
where $ \mathbf{L} = \mathbf{D} - \mathbf{G} $ is the Laplacian matrix of the graph matrix $ \mathbf{G} $ \cite{chung1997spectral} and $ \mathbf{D} $ is a diagonal matrix with $ D_{ii} = \sum_{j=1}^{T}G_{ij} $. This objective function tries to separate the transformed samples $ \mathbf{y}_{i} $ and $ \mathbf{y}_{j} $ when $ G_{ij} < 0 $ but attempts to make them close when $ G_{ij} > 0 $.

\paragraph{Constraint}
Given a stimulus, suppose $ \{ \mathbf{z}_{i} \in \mathbb{R}^{V_{i}} \}_{i=1}^{M} $ are subjects' corresponding fMRI responses and the authentic aligning maps $ \{ f_{i}:\mathbb{R}^{V_{i}}\mapsto \mathbb{R}^{K} \}_{i=1}^{M} $ are already there. Since each subject's fMRI responses to the same stimulus behave like a random variable, $ \{ f_{i}(\mathbf{z}_{i}) \}_{i=1}^{M} $ are expected to be from the same shared random variable. In other words, we do not require that $ f_{i}(\mathbf{z}_{i}) = f_{j}(\mathbf{z}_{j}) $ for any $ i,j $. Therefore, the statistical constraint $ \mathbf{Y}\mathbf{Y}^{T} = \mathbf{I} $ can be applied directly even if some samples are expected to be from the same latent response. The constraint means that each extracted shared feature is on the same scale and they are restricted be as uncorrelated as possible. The crude framework is

\begin{equation} \label{pro:NGDM}
\begin{gathered}
\argmin_{\mathbf{W}} \mathrm{tr}\left( \mathbf{W}^{T}\mathbf{\Phi}_{*}\mathbf{L}\mathbf{\Phi}_{*}^{T}\mathbf{W} \right) \\
\text{ subject to } \mathbf{W}^{T}\mathbf{\Phi}_{*}\mathbf{\Phi}_{*}^{T}\mathbf{W} = \mathbf{I} \ .
\end{gathered}	
\end{equation}

\paragraph{Relationship between GDM and Hyperalignment}
The Hyperalignment (HA) method is based on temporally-aligned dataset, assuming that $ T_{i} = T_{0} $ for $ i = 1,2,\dots,M $. Define a graph $ \mathbf{G}_{HA} $ by setting $ G_{ij} = 1/M $ when the $ i $-th and $ j $-th samples are aligned and $ G_{ij}=0 $ otherwise. Then, $ \left\lVert \mathbf{W}_{i}^{T}\mathbf{X}_{i} - \mathbf{S} \right\rVert_{F}^{2} $, which is the objective function of HA, is equal to $ \mathrm{tr}\left( \mathbf{W}^{T}\mathbf{X}_{*}\left( \mathbf{I}_{MT_{0}\times MT_{0}} - \mathbf{G}_{HA} \right)\mathbf{X}_{*}^{T}\mathbf{W} \right) $ since
\begin{gather*}
\begin{flalign*}
{} &\sum_{i=1}^{M}\| \mathbf{W}_{i}^{T}\mathbf{X}_{i} - \mathbf{S}^{*} \|_{F}^{2}\\
= & \sum_{i=1}^{M}\| \mathbf{W}_{i}^{T}\mathbf{X}_{i} \|_{F}^{2} + \left( M\| \mathbf{S}^{*} \|_{F}^{2} - 2\langle \sum_{i=1}^{M}\mathbf{W}_{i}^{T}\mathbf{X}_{i}, \mathbf{S}^{*} \rangle \right)\\
=& \mathrm{tr}\left( \mathbf{W}^{T}\mathbf{X}_{*}\mathbf{X}_{*}^{T}\mathbf{W} \right) - \mathrm{tr}\left( \mathbf{W}^{T}\mathbf{X}_{*}\mathbf{G}_{HA}\mathbf{X}_{*}^{T}\mathbf{W} \right)\\
= & \mathrm{tr}\left( \mathbf{W}^{T}\mathbf{X}_{*}\left( \mathbf{I}_{MT_{0}\times MT_{0}} - \mathbf{G}_{HA} \right)\mathbf{X}_{*}^{T}\mathbf{W} \right) \ .
\end{flalign*}
\end{gather*}
where the optimal $ \mathbf{S}^{*} $ is $ 1/M\sum_{i=1}^{M}\mathbf{W}_{i}^{T}\mathbf{X}_{i} $.

\paragraph{Computational Cost}
Problem (\ref{pro:NGDM}) is a generalized eigenvalue problem, which has been studied extensively. However, with linear kernel, the size of $ \mathbf{X}_{*}\mathbf{L}\mathbf{X}_{*}^{T} $ or $ \mathbf{X}_{*}\mathbf{X}_{*}^{T} $, which is $ \left( \sum_{i=1}^{M}V_{i} \right)^{2} $, is too tremendous to be used. For example, the dataset DS001 used in our experiment includes $ 16 $ subjects with $ 19174 $ features per subject, and then it requires at least $ 350 $ GB to store  $ \mathbf{X}_{*}\mathbf{L}\mathbf{X}_{*}^{T} $ or $ \mathbf{X}_{*}\mathbf{X}_{*}^{T} $ of shape $ (16\times 19174) \times (16\times 19174) $, which is not affordable. Thus, an efficient feasible optimization is needed. The Proposition below is helpful for solving such an issue.

\begin{proposition} \label{th:redundant}
	If $ \mathbf{W} $ is one solution for problem (\ref{pro:NGDM}), then there must be another solution that belongs to $ \mathrm{R}(\mathbf{\Phi}_{*}) $, and has the same objective value as $ \mathbf{W} $.
\end{proposition}

\emph{Proof}. $ \mathbf{W} $ can be decomposed uniquely as $ \mathbf{W} = \mathbf{W}_{R} + \mathbf{W}_{N} $ where $ \mathbf{W}_{R} \in \mathrm{R}(\mathbf{\Phi}_{*}) $ and $ \mathbf{W}_{N} \in \mathrm{N}(\mathbf{\Phi}_{*}^{T}) $. Since
\begin{gather*}
\mathbf{W}^{T}\mathbf{\Phi}_{*} = \mathbf{W}_{R}^{T}\mathbf{\Phi}_{*} + \mathbf{W}_{N}^{T}\mathbf{\Phi}_{*} = \mathbf{W}_{R}^{T}\mathbf{\Phi}_{*} + \mathbf{0} = \mathbf{W}_{R}^{T}\mathbf{\Phi}_{*} \ ,
\end{gather*}
Plugging $ \mathbf{W}_{R} $ into problem (\ref{pro:NGDM}) leads to that $ \mathbf{W}_{R} $ satisfies the constraint and shares the same objective value with $ \mathbf{W} $. \hfill $ \Box $

\paragraph{Regularized Framework} In Proposition \ref{th:redundant}, the trivial part $ \mathbf{W}_{N} $ exists due to the HSLT resolution of fMRI, i.e., $ T_{i}\ll V_{i} $. Such a trifling part indicates that it does not help produce a better solution, and thus there are many optimal solutions. If the trivial part is excluded by constraint, the optimal solution, sometimes, become unique, and a feasible optimization will be there. More details about the uniqueness are included in the \emph{Supplementary File}. In a nutshell, the regularized framework is expressed as
\begin{equation} \label{pro:GDM}
\begin{gathered}
\argmin_{\mathbf{W}} \mathrm{tr}\left( \mathbf{W}^{T}\mathbf{\Phi}_{*}\mathbf{L}\mathbf{\Phi}_{*}^{T}\mathbf{W} \right) \\
\text{ subject to } \mathbf{W}^{T}\mathbf{\Phi}_{*}\mathbf{\Phi}_{*}^{T}\mathbf{W} = \mathbf{I}\\
\mathbf{w}_{i} \in \mathrm{R}(\mathbf{\Phi}_{*}) \text{ for } 1\leq i\leq K \ .
\end{gathered}	
\end{equation}

\subsection{Optimization}
\paragraph{Naive Kernel-Based Optimization}
A simple way to solve GDM in Eq.~(\ref{pro:GDM}) is to let $ \mathbf{W}_{i} $ be $ \mathbf{X}_{i}\mathbf{B}_{i} $, where $ \mathbf{B}_{i} $ is a new variable. Then, $ \mathbf{W}^{T}\mathbf{X}_{*}\mathbf{L}\mathbf{X}_{*}^{T}\mathbf{W} $ becomes $ \mathbf{B}^{T}\mathbf{K}_{*}\mathbf{L}\mathbf{K}_{*}\mathbf{B} $, where $ \mathbf{B} $ is constructed like $ \mathbf{W} $. The optimal solution of GDM can be achieved by solving a generalized eigenvalue problem. However, with any kernel, the complexity in terms of $ \{ T_{i} \}_{i=1}^{M} $ will be at least $ O(T^{3}) $ where $ T=\sum_{i=1}^{M}T_{i} $, meaning that it heavily depends on the number of samples. In the following, we propose a more efficient kernel-based optimization algorithm.

\paragraph{Proposed Kernel-Based Optimization}
Here are some tricks to solve problem (\ref{pro:GDM}).  For each $ i $, by spectral decomposition, $ \mathbf{K}_{i} = \mathbf{V}_{i}\mathbf{D}_{i}\mathbf{V}_{i}^{T} $
where zero eigenvalues of $ \mathbf{K}_{i} $ are excluded. With $ \mathbf{U}_{i} = \mathbf{\Phi}_{i}\mathbf{V}_{i}\mathbf{D}_{i}^{-\frac{1}{2}} $, it leads to a Singular Vector Decomposition (SVD) of $ \mathbf{\Phi}_{i} $ as
\begin{equation} \label{eq:svd}
\begin{gathered}
\mathbf{\Phi}_{i} = \mathbf{U}_{i}\mathbf{D}_{i}^{\frac{1}{2}}\mathbf{V}_{i}^{T} \ .
\end{gathered}	
\end{equation}
As shown in the \emph{Supplementary File}, $ \mathbf{\Phi}_{i} $ can be decomposed similarly when the dimension of $ \mathcal{H}_{i} $ is infinite. Thus, the development below is without loss of generality. With Eq.~(\ref{eq:svd}), $ \mathbf{\Phi}_{*} = \mathbf{U}_{*}\mathbf{D}_{*}^{\frac{1}{2}}\mathbf{V}_{*}^{T} $ and then problem (\ref{pro:GDM}) is equivalent to
\begin{equation} \label{pro:tGDM}
\begin{gathered}
\argmin_{\mathbf{Q}} \mathrm{tr}\left( \mathbf{Q}^{T}\mathbf{V}_{*}^{T}\mathbf{L}\mathbf{V}_{*}\mathbf{Q} \right) \\
\text{ subject to } \mathbf{Q}^{T}\mathbf{Q} = \mathbf{I} \ .
\end{gathered}	
\end{equation}
To see this, denote the shape of $ \mathbf{D}_{*} $ by $ S\times S $. Let $ \mathcal{S} $ be $ \{ \mathbf{W}: \mathbf{W}^{T}\mathbf{\Phi}_{*}\mathbf{\Phi}_{*}^{T}\mathbf{W} = \mathbf{I} \text{ and } \mathbf{w}_{i}\in \mathrm{R}(\mathbf{\Phi}_{*}) \text{ for } 1\leq i\leq K \} $ and $ \mathcal{T} $ be $ \{ \mathbf{Q}\in \mathbb{R}^{S\times K} :\mathbf{Q}^{T}\mathbf{Q} = \mathbf{I} \} $. Denote a map $ g:\mathcal{S}\mapsto \mathcal{T} $ by setting $ g(\mathbf{W}) = \mathbf{D}_{*}^{\frac{1}{2}}\mathbf{U}_{*}^{T}\mathbf{W} $.  Since each column of $ \mathbf{W} $ belongs to $ \mathrm{R}(\mathbf{\Phi_{*}}) = \mathrm{R}(\mathbf{U}_{*}) $, $ \mathbf{U}_{*}\mathbf{D}_{*}^{-\frac{1}{2}}\mathbf{D}_{*}^{\frac{1}{2}}\mathbf{U}_{*}^{T}\mathbf{W} = \mathbf{W} $, which in turn leads to that $ g $ is a bijection between $ \mathcal{S} $ and $ g(\mathcal{S}) = \mathcal{T} $. Plugging $ \mathbf{W} = \mathbf{U}_{*}\mathbf{D}_{*}^{-\frac{1}{2}}\mathbf{Q} $ into problem (\ref{pro:GDM}) leads to problem (\ref{pro:tGDM}).

\begin{proposition}
	Using spectral decomposition, $ \mathbf{V}_{*}^{T}\mathbf{L}\mathbf{V}_{*} = \mathbf{E}\mathbf{\Lambda}\mathbf{E}^{T} $ where all eigenvalues of $ \mathbf{V}_{*}^{T}\mathbf{L}\mathbf{V}_{*} $ along the diagonal of $ \mathbf{\Lambda} $ from the top left to the bottom right are in ascending order. Denote the shape of $ \mathbf{V}_{*}^{T}\mathbf{L}\mathbf{V}_{*} $ by $ S\times S $. If $ K \leq S $,  the first $ K $ columns of $ \mathbf{E} $ is optimal for problem (\ref{pro:tGDM}).
\end{proposition}

\emph{Proof}. Firstly, problem (\ref{pro:tGDM}) is equivalent to
\begin{equation} \label{pro:sGDM}
\begin{gathered}
\argmin \mathrm{tr}\left( \mathbf{R}^{T}\mathbf{\Lambda}\mathbf{R} \right)\\
\text{ subject to } \mathbf{R}^{T}\mathbf{R} = \mathbf{I}
\end{gathered}	
\end{equation}
where $ \mathbf{R} = \mathbf{E}^{T}\mathbf{Q} $. As $ \mathbf{R}^{T}\mathbf{R} = \mathbf{I} $ infers $ \sum_{i=1}^{S}\sum_{j=1}^{K}R_{ij}^{2} = K $ and $ \sum_{j=1}^{K}R_{ij}^{2} \leq 1 $ for each $ i $, there is
\begin{gather*}
\mathrm{tr}\left( \mathbf{R}^{T}\mathbf{\Lambda}\mathbf{R} \right) = \sum_{i=1}^{S}\Lambda_{ii}\sum_{j=1}^{K}R_{ij}^{2} \geq \sum_{i=1}^{K}\Lambda_{ii} \ .
\end{gather*}

Let $ \mathbf{R}^{*} $ denote $ \begin{pmatrix}
\mathbf{I}_{K\times K} & \mathbf{0}_{K\times(S-K)}
\end{pmatrix}^{T} $. Since $ \mathrm{tr}\left( (\mathbf{R}^{*})^{T}\mathbf{\Lambda}\mathbf{R}^{*} \right) = \sum_{i=1}^{K}\Lambda_{ii} $, $ \mathbf{R}^{*} $ is optimal. Therefore, an optimal solution $ \mathbf{Q}^{*} = \mathbf{E}\mathbf{R}^{*} $ for problem (\ref{pro:tGDM}) is indeed the first $ K $ columns of $ \mathbf{E} $. \hfill $ \Box $

\paragraph{An Optimal Solution for Regularized Framework and Its Uniqueness} 
Let $ \hat{\mathbf{E}} $ denote the first $ K $ columns of $ \mathbf{E} $ and take Eq.~(\ref{eq:svd}) into consideration, then an optimal solution for problem (\ref{pro:GDM}) is
\begin{equation}
\begin{gathered}
\mathbf{W}^{*} = \mathbf{U}_{*}\mathbf{D}_{*}^{-\frac{1}{2}}\hat{\mathbf{E}} = \mathbf{\Phi}_{*}\mathbf{V}_{*}\mathbf{D}_{*}^{-1}\hat{\mathbf{E}} \ .
\end{gathered}	
\end{equation}

Since each $ \mathbf{W}_{i} $ is separable from $ \mathbf{W} $, an optimal solution for subject $ i $ is
\begin{equation} \label{eq:optimal}
\begin{gathered}
\mathbf{W}_{i}^{*} = \mathbf{\Phi}_{i}\mathbf{V}_{i}\mathbf{D}_{i}^{-1}\hat{\mathbf{E}}_{i},
\end{gathered}	
\end{equation}
where $ \{ \hat{\mathbf{E}}_{i} \}_{i=1}^{M} $ are block matrices of $ \hat{\mathbf{E}} $, which is cut along the first dimension according to the dimensions of block matrices in $ \mathbf{D}_{*} $. \

By the equivalences above, if $ K > S $, there is no solution satisfying the constraint in problem (\ref{pro:GDM}) or (\ref{pro:sGDM}) as there is no $ \mathbf{R} $ satisfying $ \mathbf{R}^{T}\mathbf{R} = \mathbf{I} $. If $ K=S $, or $ K<S $ with $ \Lambda_{KK} < \Lambda_{(K+1)(K+1)} $, the optimal solution of problem (\ref{pro:GDM}) is unique except being rotated. In other words, if $ \mathbf{W}^{(1)} $ and $ \mathbf{W}^{(2)} $ are two optimal solutions, there is an orthogonal matrix $ \mathbf{P} $ such that $ \mathbf{W}^{(1)} = \mathbf{W}^{(2)}\mathbf{P} $. By the definition of $ \mathbf{W} $, it implies that the shared feature space is unique except being rotated. More details are given in the \emph{Supplementary File}.

\subsection{Low-Dimension Assumption}
\paragraph{Potential Overfitting of GDM} Suppose the dataset $ \{ \mathbf{X}_{i} \}_{i=1}^{M} $ is temporally-aligned, which means that $ T_{i} = T_{j} = T_{0} $ for any $ i,j $. Construct a graph matrix $ \mathbf{G} $ by setting $ G_{ij} = 1 $ if the $ i $-th and $ j $-th samples are temporally-aligned, and $ G_{ij} = 0 $ otherwise. With this graph matrix, the objective function of problem (\ref{pro:GDM}) with linear kernel becomes
\begin{gather}
\argmin_{\mathbf{W}_{i}} \cfrac{1}{2}\sum_{i=1}^{M}\sum_{j=1}^{M}\left\lVert \mathbf{W}_{i}^{T}\mathbf{X}_{i} - \mathbf{W}_{j}^{T}\mathbf{X}_{j} \right\rVert_{F}^{2} \ .
\end{gather}

Assume that each $ \mathbf{X}_{i} \in \mathbb{R}^{V_{i}\times T_{0}} $ is full-column rank. Let $ \mathbf{P}_{K\times T_{0}} $ ($ K\leq T_{0} $) be any matrix such that $ \mathbf{P}\mathbf{P}^{T} = \mathbf{I} $ and take Eq.~(\ref{eq:svd}) into consideration where $ \mathbf{\Phi}_{i} $ is replaced by $ \mathbf{X}_{i} $. With $ \mathbf{W}^{*}_{i} = M^{-1}\mathbf{U}_{i}\mathbf{D}_{i}^{-\frac{1}{2}}\mathbf{V}_{i}^{T}\mathbf{P}^{T} $ and $ (\mathbf{W}^{*})^{T} = \begin{pmatrix}
(\mathbf{W}_{1}^{*})^{T} & \cdots & (\mathbf{W}_{M}^{*})^{T}
\end{pmatrix} $, $ \mathbf{W}^{*} $ satisfies the constraints in problem (\ref{pro:GDM}). However, $ (\mathbf{W}^{*}_{i})^{T}\mathbf{\Phi}_{i} = M^{-1}\mathbf{P} $ for each $ i $, which implies that the generated optimal solution (\ref{eq:optimal}) aligns each aligning sample perfectly. The culprit is the full-column rank assumption of each $ \mathbf{X}_{i} $, which is almost the case due to the HSLT resolution of fMRI, i.e., $ T_{0} \ll V_{i} $. Therefore, we impose a low-dimension assumption over each new feature space $ \mathcal{H}_{i} $, which conforms with that the number of informative features is usually much less than the number of voxels. Suppose the low-dimension in $ \mathcal{H}_{i} $ is $ L_{i} $, then we try to fit the data in $ \mathcal{H}_{i} $ by an $ L_{i} $ dimensional affine subspace\footnote{An $ L $ dimensional affine subspace in $ \mathbb{R}^{N} $ is $ \mathcal{V}+\mathbf{c} $ where $ \mathcal{V} $ is an $ L $ dimensional subspace and $ \mathbf{c} \in \mathbb{R}^{N} $.}, i.e.,
\begin{equation}
\begin{gathered}
\argmin_{\substack{\mathbf{m}_{i}\in \mathbb{R}^{N_{i}} \\ \mathbf{F}_{i}\in\mathbb{R}^{N_{i}\times L_{i}}}} \sum_{j=1}^{T_{i}}\left\lVert \mathbf{F}_{i}\mathbf{F}_{i}^{T}((\boldgreek{\phi}_{i})_{j}-\mathbf{m}_{i})-((\boldgreek{\phi}_{i})_{j}-\mathbf{m}_{i}) \right\rVert^{2}_{F}\\
\text{ subject to } \mathbf{F}_{i}^{T}\mathbf{F}_{i}= \mathbf{I} \ .
\end{gathered}	
\end{equation}

An optimal solution is $ \mathbf{m}_{i}^{*} = T_{i}^{-1}\sum_{j=1}^{T_{i}}(\boldgreek{\phi}_{i})_{j} $ and $ \mathbf{F}_{i}^{*} $ be the first $ L_{i} $ columns of $ \mathbf{U}_{i} $ in Eq.~(\ref{eq:svd}) where $ (\boldgreek{\phi}_{i})_{j} \gets (\boldgreek{\phi}_{i})_{j}-\mathbf{m}_{i}^{*} $ for $ 1\leq j\leq T_{i} $. The general proof for any Hilbert space is left in the \emph{Supplementary File}.

\paragraph{Centralizing over Gram Matrices} To generate and apply $ \mathbf{F}_{i}^{*} $, it is necessary to centralize all data by the mean of the aligning data, i.e., $ (\boldgreek{\phi}_{i})_{j} \gets (\boldgreek{\phi}_{i})_{j}-\mathbf{m}_{i}^{*} $. Suppose $ \mathbf{Z}_{i}\in \mathbb{R}^{V_{i}\times E_{i}} $ is extra fMRI data for the $ i $-th subject. Denote all-one matrices by $ \mathbf{J} $. For subject $ i $, the centralizing can be applied on the Gram matrices directly since
\begin{equation}
\begin{gathered}
\left( \Phi_{i}(\mathbf{Z}_{i})^{T} - T_{i}^{-1}\mathbf{J}_{E_{i}\times T_{i}}\mathbf{\Phi}_{i}^{T} \right)\left( \mathbf{\Phi}_{i} - T_{i}^{-1}\mathbf{\Phi}_{i}\mathbf{J}_{T_{i}\times T_{i}} \right)\\ = \Phi_{i}(\mathbf{Z}_{i})^{T}\mathbf{\Phi}_{i} + T_{i}^{-2}\mathbf{J}_{E_{i}\times T_{i}}\mathbf{\Phi}_{i}^{T}\mathbf{\Phi}_{i}\mathbf{J}_{T_{i}\times T_{i}} \\ - T_{i}^{-1}\mathbf{J}_{E_{i}\times T_{i}}\mathbf{\Phi}_{i}^{T}\mathbf{\Phi}_{i} - T_{i}^{-1}\Phi_{i}(\mathbf{Z}_{i})^{T}\mathbf{\Phi}_{i}\mathbf{J}_{T_{i}\times T_{i}} \ .
\end{gathered}	
\end{equation}

From now on, suppose all Gram matrices have been centralized. As is provided in Eq.~(\ref{eq:svd}), $ \mathbf{\Phi}_{i} = \mathbf{V}_{i}\mathbf{D}_{i}^{\frac{1}{2}}\mathbf{U}_{i}^{T} $, which is an SVD. Denote the number of the (non-zero) singular values in $ \mathbf{D}_{i} $ by $ s_{i} $. Assume the singular values in $ \mathbf{D}_{i}^{\frac{1}{2}} $ are in descending order and the first $ L_{i}  $ $ (L_{i} \leq s_{i}) $ singular values approximately contains $ p_{i}\% $ ($ p_{i} \in (0,100] $) energy , i.e., $ \sum_{j=1}^{L_{i}}(D_{i})_{jj}^{\frac{1}{2}} / \sum_{j=1}^{s_{i}}(D_{i})_{jj}^{\frac{1}{2}}  \approx p_{i}\% $. By this way, the low dimension $ L_{i} $ is controlled by $ p_{i}\% $. Therefore, the corresponding low-dimensional representation of $ \Phi_{i}(\mathbf{Z}_{i}) $ would be $ \hat{\mathbf{U}}_{i}\hat{\mathbf{U}}_{i}^{T}\Phi_{i}(\mathbf{Z}_{i}) $ where $ \hat{\mathbf{U}}_{i} $ is the first $ L_{i} $ columns of $ \mathbf{U}_{i} $. Generally, with only Gram matrices, there is

\begin{equation*}
\begin{gathered}
\Phi_{i}(\mathbf{Z}_{i})^{T}\hat{\mathbf{U}}_{i}\hat{\mathbf{U}}_{i}^{T}\hat{\mathbf{U}}_{i}\hat{\mathbf{U}}_{i}^{T}\mathbf{\Phi}_{i} = \Phi_{i}(\mathbf{Z}_{i})^{T}\hat{\mathbf{U}}_{i}\hat{\mathbf{U}}_{i}^{T}\mathbf{\Phi}_{i}\\
\not= \Phi_{i}(\mathbf{Z}_{i})^{T}\mathbf{\Phi}_{i} \ .
\end{gathered}	
\end{equation*}
Nevertheless, the equality holds with the help of $ \hat{\mathbf{V}}_{i} $ that is defined by the first $ L_{i} $ columns of $ \mathbf{V}_{i} $

\begin{proposition}
	\begin{equation} \label{eq:pca}
	\begin{gathered}
	\Phi_{i}(\mathbf{Z}_{i})^{T}\mathbf{\Phi}_{i}\hat{\mathbf{V}}_{i} = \Phi_{i}(\mathbf{Z}_{i})^{T}\hat{\mathbf{U}}_{i}\hat{\mathbf{U}}_{i}^{T}\mathbf{\Phi}_{i}\hat{\mathbf{V}}_{i} \ .
	\end{gathered}	
	\end{equation}
\end{proposition}

\emph{Proof}. Since $ \mathbf{\Phi}_{i}\hat{\mathbf{V}}_{i} = \mathbf{U}_{i}\mathbf{D}_{i}^{\frac{1}{2}}\mathbf{V}_{i}^{T}\hat{\mathbf{V}}_{i} = \hat{\mathbf{U}}_{i}\mathbf{\Lambda}_{i} $
where $ \mathbf{\Lambda}_{i} $ is the upper left $ L_{i}\times L_{i} $ submatrix of $ \mathbf{D}_{i}^{\frac{1}{2}} $, there is $ \hat{\mathbf{U}}_{i}\hat{\mathbf{U}}_{i}^{T}\mathbf{\Phi}_{i}\hat{\mathbf{V}}_{i} = \hat{\mathbf{U}}_{i}\hat{\mathbf{U}}_{i}^{T}\hat{\mathbf{U}}_{i}\mathbf{\Lambda}_{i} = \hat{\mathbf{U}}_{i}\mathbf{\Lambda}_{i} = \mathbf{\Phi}_{i}\hat{\mathbf{V}}_{i}  $.
\hfill $ \Box $

Therefore, the proposed kernel-based optimization can easily incorporate the low-dimension assumption over each new feature space. It will be shown in our experiments that this is essential for getting useful results. The overall optimization procedure of GDM is summarized in Algorithm \ref{alg:GDM}.

\subsection{Complexity Analysis} 
The shape of $ \hat{\mathbf{V}}_{*}^{T}\mathbf{L}\hat{\mathbf{V}}_{*} $ is $ L\times L $ where $ L = \sum_{i=1}^{M}L_{i} $ and $ L_{i} $ is the low-dimension of the $ i $-th subject. Suppose the Gram matrices are given, and $ K < T_{i} $ for each $ i $, i.e., the number of the shared features is smaller than the sample size, the complexity of GDM is $ O((\sum_{i=1}^{M}T_{i}^{3}) + L(L^{2}+T^{2}+LT)) $ where $ T = \sum_{i=1}^{M}T_{i} $. If each low-dimension $ L_{i} $ is fixed, the complexity thus becomes $ O((\sum_{i=1}^{M}T_{i}^{3})+T^{2}) $. Notably, it could be reduced into $ O(\max_{1\leq i\leq M}T_{i}^{3}+T^{2}) $ by using parallel programming. By contrast, the naive kernel scheme cannot be parallelized, and the complexity of employing it is $ O(T^{3}) $. Therefore, our proposed kernel method is more efficient compared to the naive kernel scheme. Besides, different from methods based on iterative optimization algorithms, one can obtain the optimal solution of GDM directly.

\begin{algorithm}[t] 
	\caption{Graph-Based Decoding Model (GDM)}
	\label{alg:GDM}
	\begin{algorithmic}[1]
		\Statex \textbf{Input:} Aligning data $ \{ \mathbf{X}_{i} \in \mathbb{R}^{V_{i}\times T_{i}} \}_{i=1}^{M} $, the number of the shared features $ K $, the energy $ \{ p_{i}\% \}_{i=1}^{M} $ to be kept, a specific Laplacian matrix $ \mathbf{L} $ and kernel functions for each subject.
		\State For each $ i $, standardize $ \mathbf{X}_{i} $ such that it has zero mean along the second dimension and the variance of each feature, i.e., voxel, is $ 1 $.
		\State Generate $ \{ \mathbf{K}_{i} \}_{i=1}^{M} $ via specified kernel functions.
		\State Centralize Gram matrices: $ \mathbf{K}_{i} \gets \mathbf{K}_{i} + T_{i}^{-2}\mathbf{J}_{T_{i}\times T_{i}}\mathbf{K}_{i}\mathbf{J}_{T_{i}\times T_{i}}-T_{i}^{-1}\mathbf{J}_{T_{i}\times T_{i}}\mathbf{K}_{i} - T_{i}^{-1}\mathbf{K}_{i}\mathbf{J}_{T_{i}\times T_{i}}  $
		\For{$ i \gets 1 $ \textbf{to} $ M $} \label{stp:begin}
		\State $ \mathbf{K}_{i} = \mathbf{V}_{i}\mathbf{D}_{i}\mathbf{V}_{i}^{T} $ by spectral decomposition. The eigenvalues in $ \mathbf{D}_{i} $ is in descending order.
		\State Find $ L_{i} $ such that the first $ L_{i} $ diagonal elements of $ \mathbf{D}_{i}^{\frac{1}{2}} $ contains approximately $ p_{i}\% $ energy.
		\State Let $ \hat{\mathbf{V}}_{i} $ be the first $ L_{i} $ columns of $ \mathbf{V}_{i} $\,.
		\State Let $ \hat{\mathbf{D}}_{i} $ be the top left $ L_{i} \times L_{i} $ submatrix of $ \mathbf{D}_{i} $\,.
		\EndFor \label{stp:end}
		\State By spectral decomposition, $ \hat{\mathbf{V}}_{*}^{T}\mathbf{L}\hat{\mathbf{V}}_{*} = \mathbf{E}\mathbf{\Sigma}\mathbf{E}^{T} $ where the diagonal elements of $ \mathbf{\Sigma} $ is ascending.
		\State Let $ \hat{\mathbf{E}} $ be the first $ K $ columns of $ \mathbf{E} $ and then cut $ \hat{\mathbf{E}} $ along the first dimension such that $ \hat{\mathbf{E}}_{i}\in \mathbb{R}^{L_{i}\times K} $.
		\State For $ 1\leq i\leq M $, $ \mathbf{W}_{i}^{*} \gets \mathbf{\Phi}_{i}\hat{\mathbf{V}}_{i}\hat{\mathbf{D}}_{i}^{-1}\hat{\mathbf{E}}_{i} $\,.
	\end{algorithmic}
\end{algorithm}

\section{Experiments}

\begin{table*}[t]
	\centering
	\caption{The brief information and parameter settings for each dataset. Here, $ K $ is the number of the shared features, energy $ p\% $ is set for all subjects, $ \nu $ is related to $ \nu$-SVM.}
	\label{tab:setting}
	\scriptsize
	\begin{tabular}{ccccccccc} \toprule
		\textbf{Dataset} & \#subject & \#sample/subject & \#feature & \#category & $ K $ & energy($ p\% $) & $ \nu $ & \#subject left out\\ \midrule
		\textbf{DS105WB} & 6 & 994 & 19174 & 8 & 10 & 82 & 0.8 & 1\\
		\textbf{DS105ROI} & 6 & 994 & 2294 & 8 & 10 & 82 & 0.8 & 1\\
		\textbf{DS011} & 14 & 271 & 19174 & 2 & 10 & 82 & 0.3 & 2\\
		\textbf{DS001} & 16 & 485 & 19174 & 4 & 10 & 82 & 0.5 & 4\\
		\textbf{DS232} & 10 & 1691 & 9947 & 4 & 10 & 82 & 0.8 & 2\\
		\textbf{Raider.Movie} & 10 & 2203 & 1000 & --- & 20 & 35 & --- & ---\\
		\textbf{Raider.Image} & 10 & 56 & 1000 & 7 & --- & --- & 0.5 & 2 \\ \bottomrule
	\end{tabular}	
\end{table*}

\begin{table*}[t]
	\caption{The performance on temporally-aligned datasets is measured by BSC accuracy. The larger the better. Each performance is reported by averaging accuracies over all folds with standard deviation. The bold denotes the best result on each dataset.}		
	\label{tab:result}
	\centering
	\scriptsize
	\begin{tabular}{ccccccccc} \toprule
		\textbf{Dataset(\#class)} & $ \mathbf{\nu} $\textbf{-SVM } & \textbf{HA} & \textbf{KHA} & \textbf{SVDHA} & \textbf{SRM} & \textbf{RSRM} & \textbf{RHA} & \textbf{GDM (Ours)} \\ \midrule
		\textbf{DS105WB(8)} & \begin{tabular}{@{}c@{}}$ 11.67   \pm1.80 $\end{tabular} &\begin{tabular}{@{}c@{}}$ 39.70 \pm3.90 $\end{tabular} & \begin{tabular}{@{}c@{}}$ 39.22   \pm4.50 $\end{tabular} & \begin{tabular}{@{}c@{}}$ 30.48   \pm3.52 $\end{tabular} & \begin{tabular}{@{}c@{}}$ 39.69   \pm3.95 $\end{tabular} & \begin{tabular}{@{}c@{}}$ 40.01   \pm3.84 $\end{tabular} & \begin{tabular}{@{}c@{}}$ 52.50   \pm4.28 $\end{tabular} & \begin{tabular}{@{}c@{}}  $ \mathbf{60.68}   \mathbf{\pm5.23} $\end{tabular} \\ 
		\textbf{DS105ROI(8)} & \begin{tabular}{@{}c@{}}$ 13.06   \pm2.93 $\end{tabular} & \begin{tabular}{@{}c@{}}$ 48.05   \pm3.93 $\end{tabular} & \begin{tabular}{@{}c@{}}$ 48.22   \pm3.34 $\end{tabular} & \begin{tabular}{@{}c@{}}$ 41.33   \pm4.19 $\end{tabular} & \begin{tabular}{@{}c@{}}$ 48.14   \pm3.17 $\end{tabular} & \begin{tabular}{@{}c@{}}$ 48.51   \pm3.80 $\end{tabular} & \begin{tabular}{@{}c@{}}$ 57.63   \pm5.55 $\end{tabular} & \begin{tabular}{@{}c@{}}$ \mathbf{62.22}   \mathbf{\pm4.23} $\end{tabular} \\ 
		\textbf{DS011(2)} & \begin{tabular}{@{}c@{}}$ 51.80   \pm3.73 $\end{tabular} & \begin{tabular}{@{}c@{}}$ 85.39   \pm3.52 $\end{tabular} & \begin{tabular}{@{}c@{}}$ 85.79   \pm3.82 $\end{tabular} & \begin{tabular}{@{}c@{}}$ 74.42   \pm4.40 $\end{tabular} & \begin{tabular}{@{}c@{}}$ 85.47   \pm3.53 $\end{tabular} & \begin{tabular}{@{}c@{}}$ 85.58   \pm3.89 $\end{tabular} & \begin{tabular}{@{}c@{}}$ 91.80   \pm2.65 $\end{tabular} & \begin{tabular}{@{}c@{}}$ \mathbf{92.49}   \mathbf{\pm2.24} $\end{tabular} \\ 
		\textbf{DS232(4)} & \begin{tabular}{@{}c@{}}$ 25.89   \pm2.46 $\end{tabular} & \begin{tabular}{@{}c@{}}$ 69.34   \pm3.22 $\end{tabular} & \begin{tabular}{@{}c@{}}$ 69.38   \pm3.16 $\end{tabular} & \begin{tabular}{@{}c@{}}$ 56.77   \pm4.52 $\end{tabular} & \begin{tabular}{@{}c@{}}$ 69.18   \pm3.27 $\end{tabular} & \begin{tabular}{@{}c@{}}$ 69.25   \pm3.20 $\end{tabular} & \begin{tabular}{@{}c@{}}$ 77.64   \pm2.75 $\end{tabular}  & \begin{tabular}{@{}c@{}}$ \mathbf{82.47}   \mathbf{\pm1.45} $\end{tabular} \\ 
		\textbf{DS001(4)} & \begin{tabular}{@{}c@{}}$ 34.32   \pm2.08 $\end{tabular} & \begin{tabular}{@{}c@{}}$ 56.74   \pm1.63 $\end{tabular} & \begin{tabular}{@{}c@{}}$ 57.10   \pm1.97 $\end{tabular} & \begin{tabular}{@{}c@{}}$ 51.99   \pm1.87 $\end{tabular} & \begin{tabular}{@{}c@{}}$ 56.83   \pm1.54 $\end{tabular} & \begin{tabular}{@{}c@{}}$ 57.20   \pm1.30 $\end{tabular} & \begin{tabular}{@{}c@{}}$ 57.87   \pm0.61 $\end{tabular} & \begin{tabular}{@{}c@{}}$ \mathbf{62.68}   \mathbf{\pm1.53} $\end{tabular} \\ 
		\textbf{Raider(7)} & \begin{tabular}{@{}c@{}}$ 26.61   \pm3.80 $\end{tabular} & \begin{tabular}{@{}c@{}}$ 60.48   \pm3.68 $\end{tabular} & \begin{tabular}{@{}c@{}}$ 60.71   \pm3.23 $\end{tabular} & \begin{tabular}{@{}c@{}}$ 58.99   \pm4.19 $\end{tabular} & \begin{tabular}{@{}c@{}}$ 60.65   \pm4.16 $\end{tabular} & \begin{tabular}{@{}c@{}}$ 62.38   \pm3.48 $\end{tabular} & \begin{tabular}{@{}c@{}}$ 59.82   \pm4.10 $\end{tabular} & \begin{tabular}{@{}c@{}}$ \mathbf{64.52}   \mathbf{\pm3.28} $\end{tabular} \\ \bottomrule	
	\end{tabular}

\end{table*}

\begin{figure*}[t]
	\begin{subfigure}{.33\textwidth}  
		\centering
		\includegraphics[width=\linewidth]{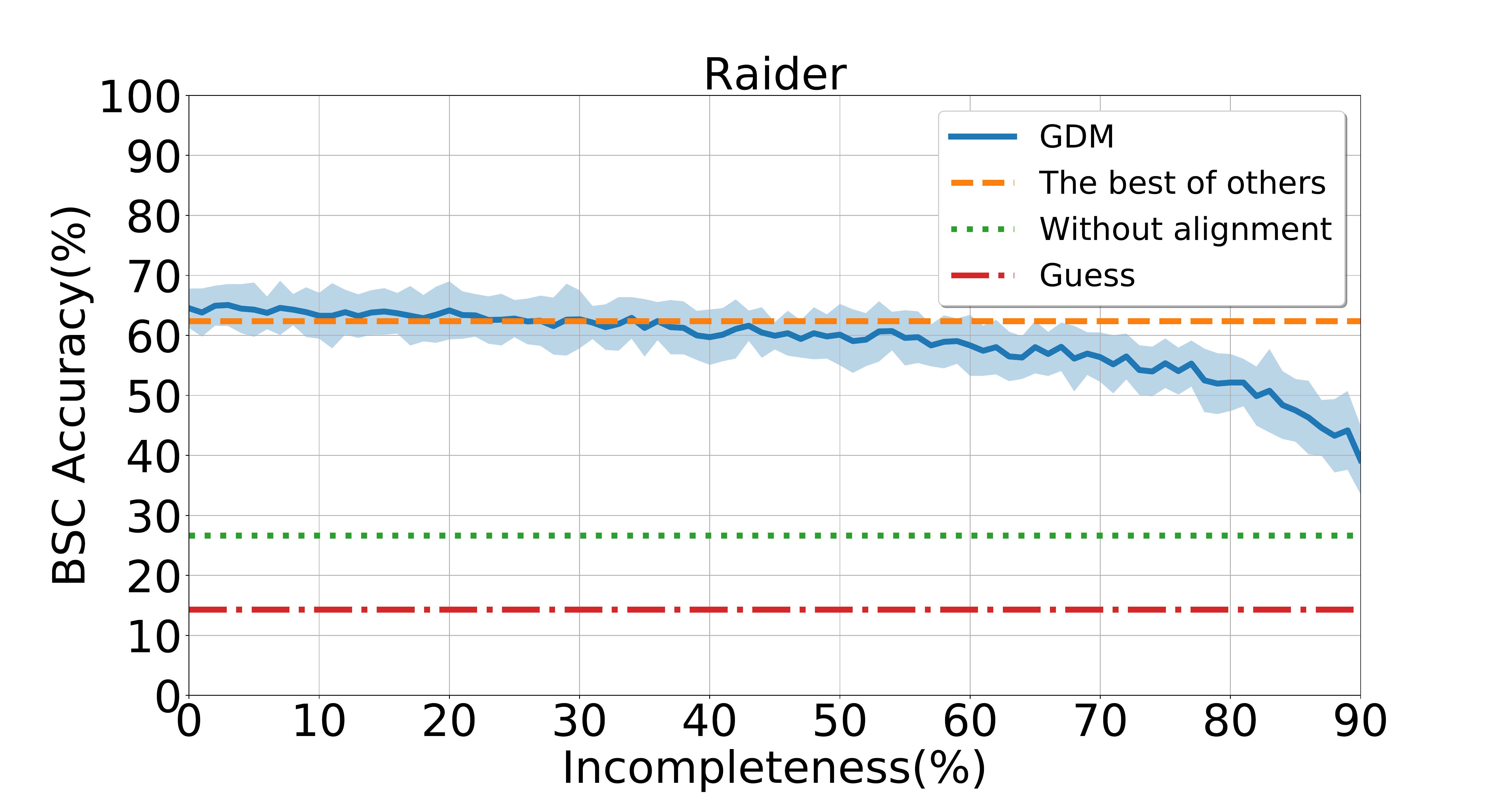}
	\end{subfigure}%
	\begin{subfigure}{.33\textwidth}
		\centering
		\includegraphics[width=\linewidth]{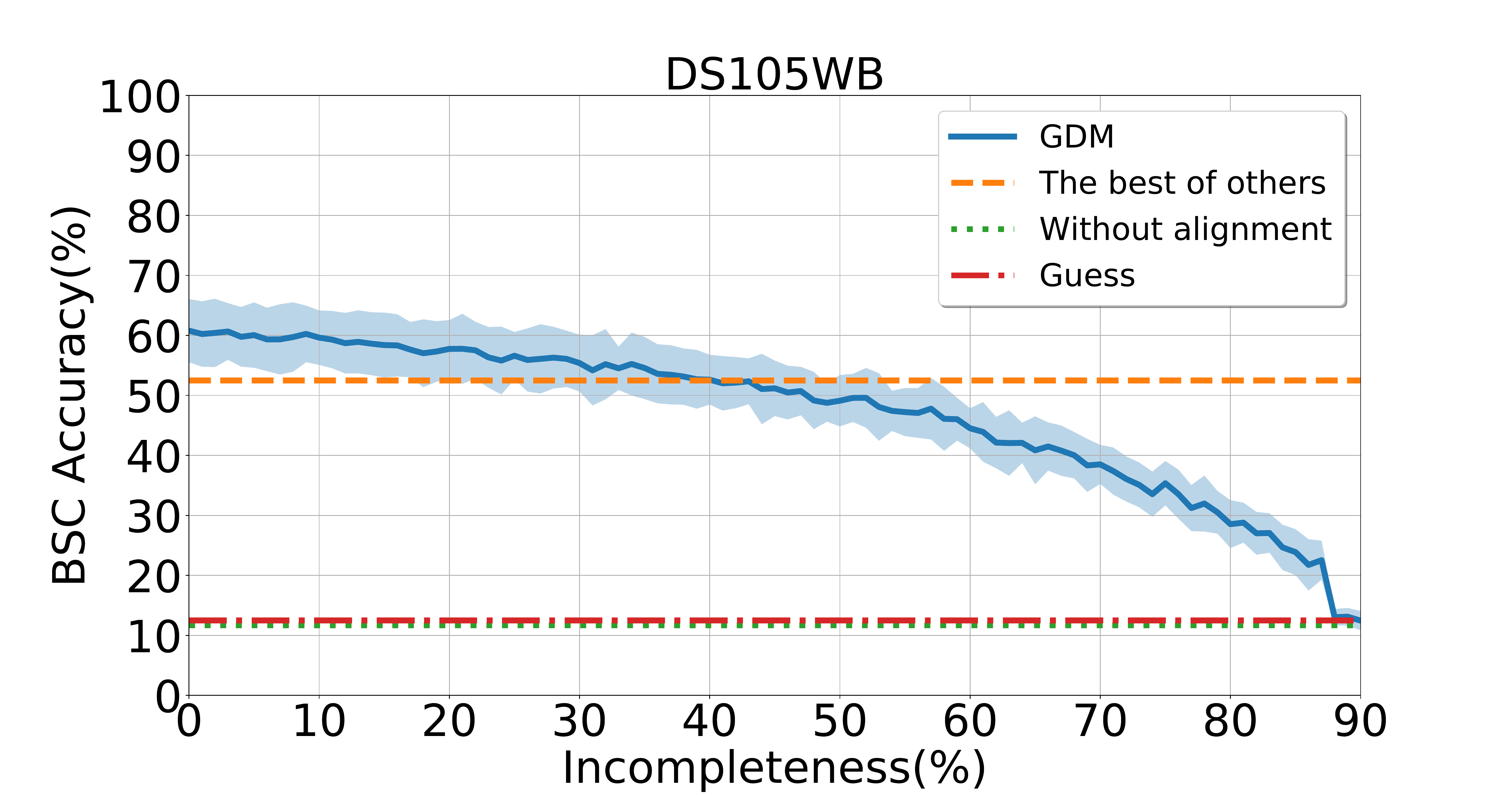}
	\end{subfigure}
	\begin{subfigure}{.33\textwidth}
		\centering
		\includegraphics[width=\linewidth]{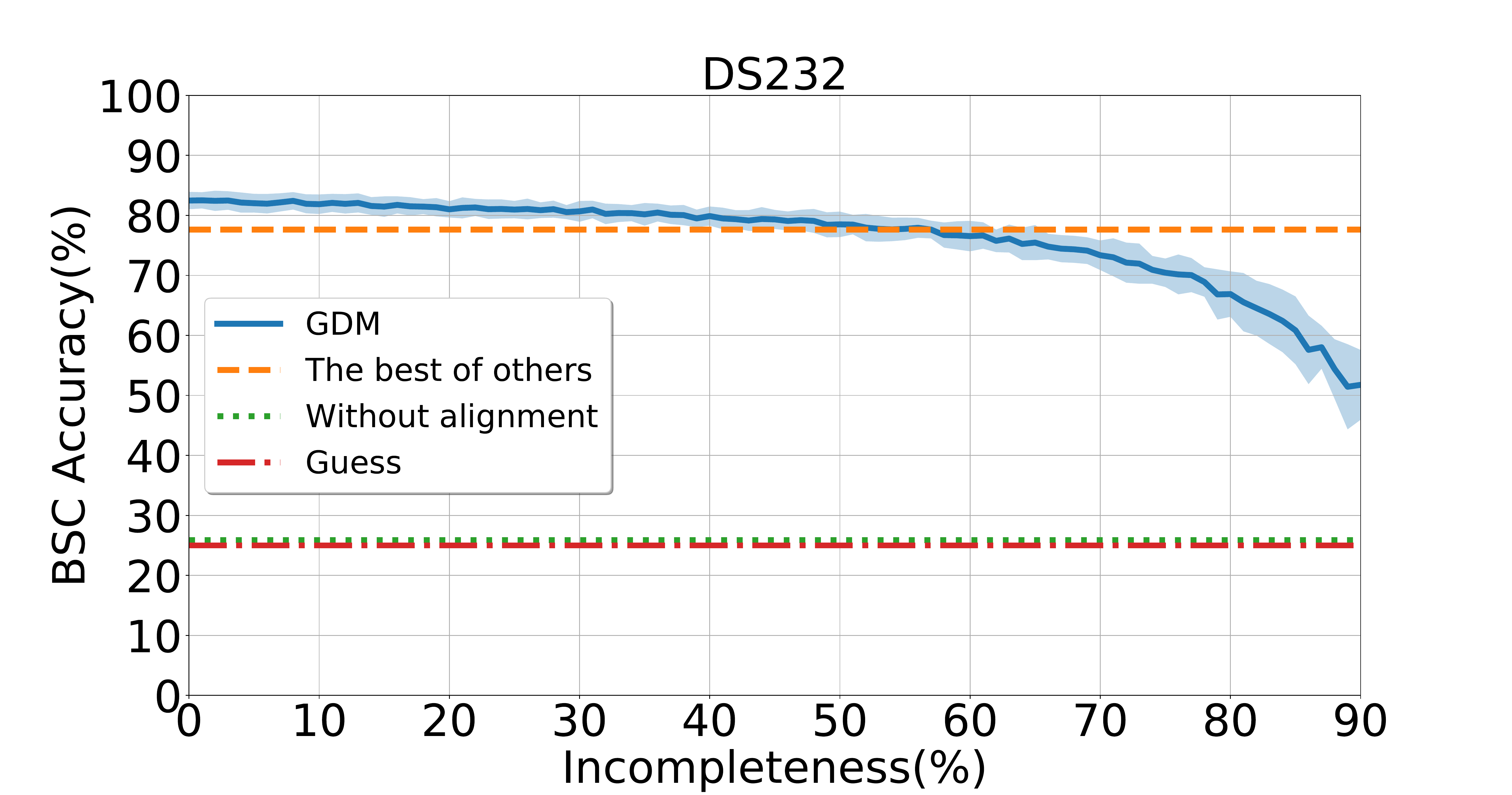}
	\end{subfigure}
	\caption{Performance of GDM on incomplete or unaligned datasets. Here, $ q\% $ incompleteness means that $ q\% $ of the aligning data are randomly removed per subject. The term ``Without alignment'' denotes the  $ \nu $-SVM method without any alignment. ``The best of others'' refers to the best result of six competing methods with complete data, and ``Guess'' denotes the randomly guess method.}
	\label{fig:incomplete}
\end{figure*}

\begin{figure*}[t]
	\begin{subfigure}{.33\textwidth}  
		\centering
		\includegraphics[width=\linewidth]{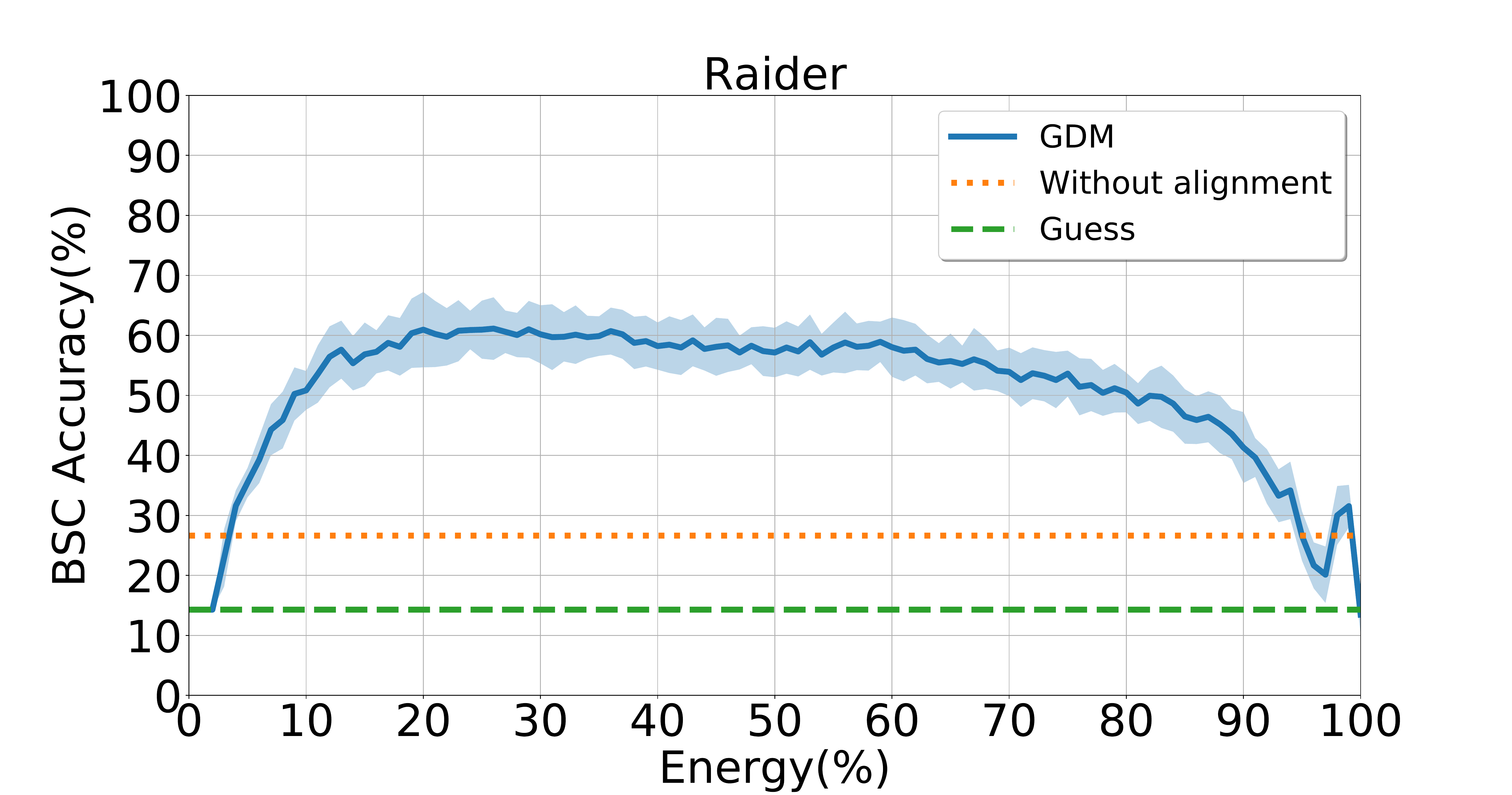}
	\end{subfigure}%
	\begin{subfigure}{.33\textwidth}
		\centering
		\includegraphics[width=\linewidth]{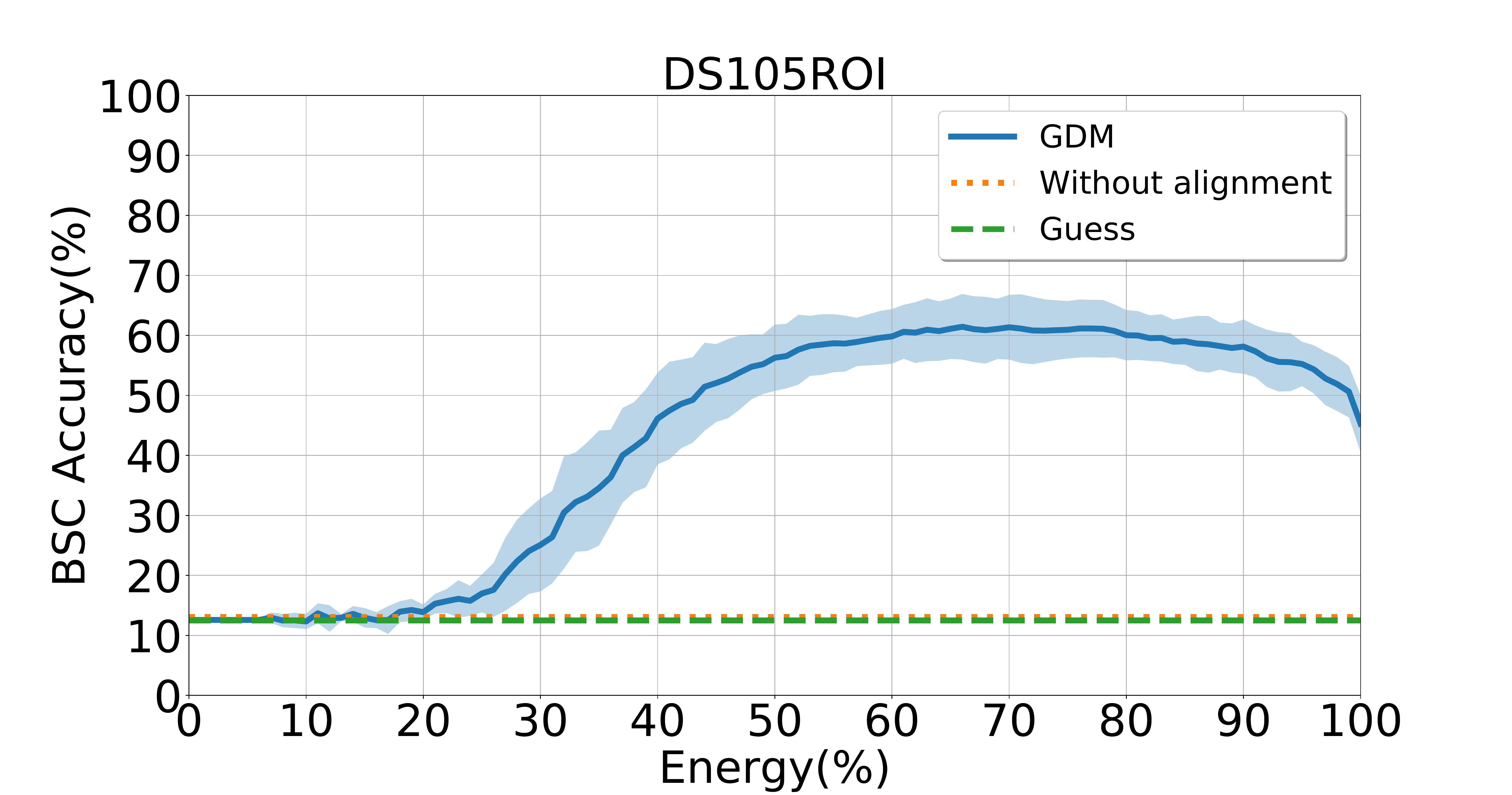}
	\end{subfigure}
	\begin{subfigure}{.33\textwidth}
		\centering
		\includegraphics[width=\linewidth]{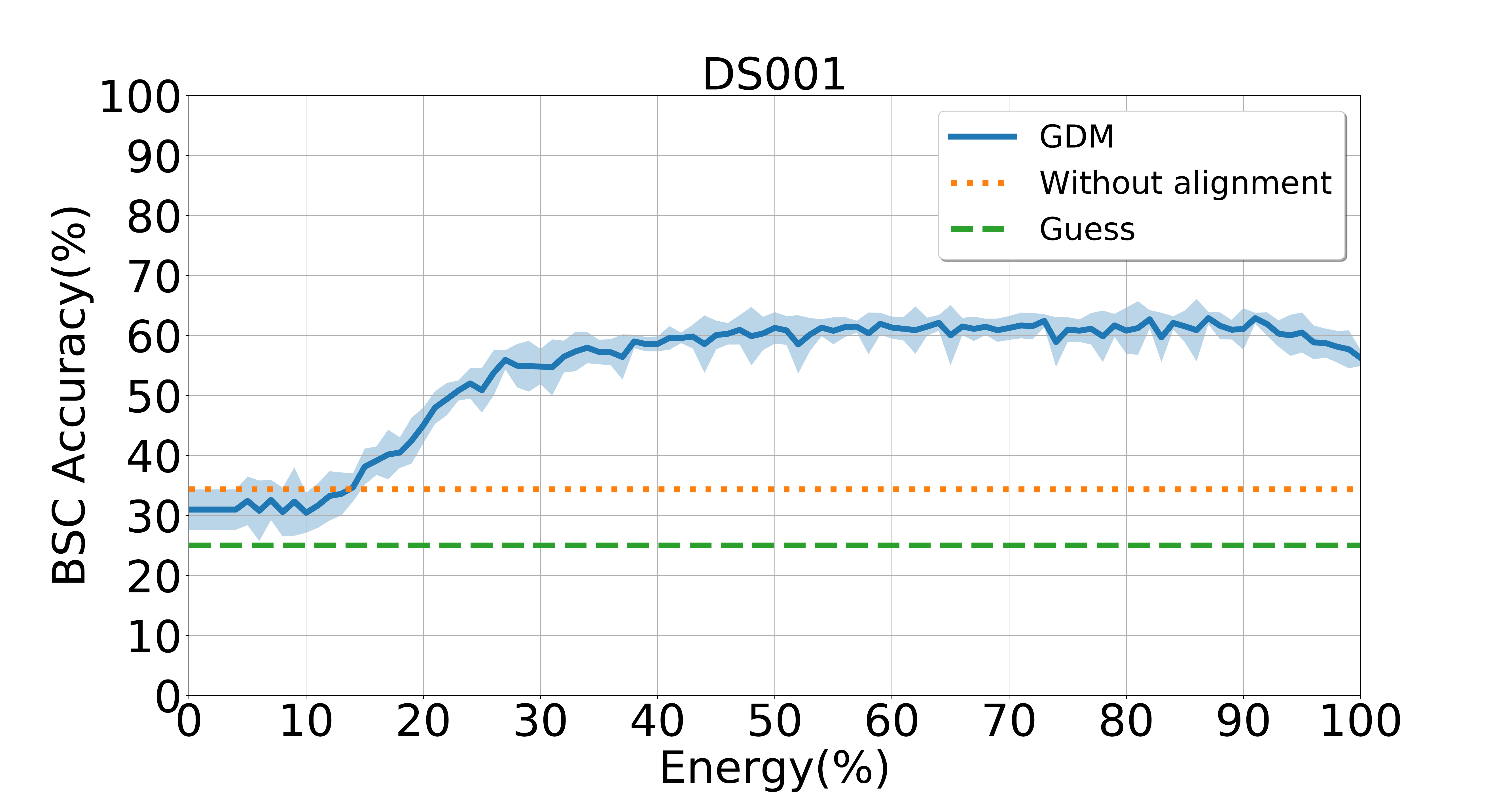}
	\end{subfigure}
	\caption{The necessity of low-dimension assumption for GDM. Here, $ p\% $ energy shows how much energy is nearly kept per subject. The term ``Without alignment'' denotes the  $ \nu $-SVM method without any alignment.}
	\label{fig:energy}
\end{figure*}

\paragraph{Materials}
We utilize five datasets shared by \url{openfmri.org} and Chen \textit{et al.}\ \cite{chen2015reduced}. The relevant information about each dataset is outlined in Table \ref{tab:setting}. Raw datasets are preprocessed by using FSL (\url{https://fsl.fmrib.ox.ac.uk/}), following a standard process (i.e., slice timing, anatomical alignment, normalization, and smoothing). The default parameters in FSL were taken when the dataset does not provide. The description of each dataset is as follows:
\begin{enumerate}[wide=10pt]
	\item[(1) \textbf{DS105}:] The fMRI data were measured while six subjects viewed gray-scale images of faces, houses, cats, bottles, scissors, shoes, chairs, and nonsense images \cite{haxby2001distributed}. Hence, there are totally 8 categories in this dataset. Here, DS105WB contains the whole-brain fMRI data while the data in DS105ROI are based on a region of interest.

	\item[(2) \textbf{DS011}:] Fourteen subjects participated in a single task (weather prediction). In the first phase, they learned to predict weather outcomes (rain or sun) for two different cities. After learning, they predicted weather \cite{foerde2006modulation}. Thus, there are two cognitive states.

	\item[(3) \textbf{DS001:}] Sixteen subjects were instructed to inflate a control balloon or a reward balloon on a screen. For a control balloon, subjects had merely one choice whereas they could choose to pump or cash out for another case. After opting to pump, the balloon may explode or expand \cite{schonberg2012decreasing}. Hence, there are four different cognitive states. 
	
	\item[(4) \textbf{DS232}:] Ten subjects were instructed to respond to images of faces, scenes, objects and phrase-scrambled versions of the scene images \cite{carlin2017adjudicating}.

	\item[(5) \textbf{Raider}:] As a commonly-used one, it collected data from 10 subjects participating in two experiments. Firstly, 10 subjects watched a movie Raiders of the Lost Ark (2203 TRs). The data of movie dataset does not contain any label. In the next experiment, the same 10 subjects were shown 7 classes of images (female face, male face, monkey face, dog face, house, chair and shoes)~\cite{chen2015reduced}.
	
\end{enumerate}

It's worth noting that, except for Raider, all four other datasets are not temporally-aligned. To compare GDM with other temporal-alignment-based methods, following the previous study in~\cite{chen2015reduced}, these datasets (i.e., DS105, DS011, DS001, and DS232) are reordered and truncated, or downsampled, to be aligned according to their categories.

\paragraph{Experimental Setup}
We follow the experiment setup with a cross-validation strategy in previous studies \cite{chen2014joint,haxby2011common}, as illustrated by Fig.~S1 and Fig.~S2 in the \emph{Supplementary File}. Specifically, except for the Raider, each subject's data is equally divided into two parts with each category being equally split, one is for alignment whereas the other is for training or testing a classifier. Switching the roles of the two parts and leave-$ k $-subject-out strategy are adopted for cross-validation. For instance, if there are $ 16 $ subjects, leave-$ 4 $-subject-out leads to $ 16\div4\times2 = 8 $ folds for cross-validation.
For Raider dataset, the movie data is taken for alignment while the image data is for classification. Here, the first $2,202$ time points of movie data are used for alignment. Then it is equally divided into threes parts with each part having 734 samples for cross-validation. Since a leave-$2$-subject-out is used in this dataset, there are a total of $ 10\div2\times3=15 $ folds when using Raider. 
As shown in Fig.~S1 in the \emph{Supplementary File}, the experiment on each dataset contains two stages: 1) aligning phase, and 2) classification phase. At the first phase, one part of all subjects' data are fed into a functional alignment method to yield the corresponding aligning maps $ \{ f_{i}:\mathbb{R}^{V}_{i}\mapsto \mathbb{R}^{K} \}_{i=1}^{M} $. At the classification phase, the remaining part of data are first mapped to the shared feature space via the learned aligning maps and then used for classification model construction. Note that those data used at the aligning phase will not be used at the classification phase in our experiments.

Since each dataset (or part of it) used in this paper includes labels, the performance of alignment is assessed by testing how well a trained classifier can generalize to new subjects, i.e., between-subject classification (BSC) accuracy \cite{haxby2011common}. Like previous studies, $ \nu $-SVM is used for classification \cite{chang2011libsvm}.

\paragraph{Competing Methods}
The proposed GDM method is compared with six state-of-the-art methods in the experiments, including (1) Hyperalignment (HA)~\cite{haxby2011common}, (2) Regularized Hyperalignment (RHA)~\cite{xu2012regularized}, (3) Kernel Hyperalignment (KHA)~\cite{lorbert2012kernel}, (4) SVD-Hyperalignment (SVDHA)~\cite{chen2014joint}, (5) Shared Response Model (SRM)~\cite{chen2015reduced}, and (6) Robust SRM (RSRM)~\cite{turek2018capturing}. All methods are implemented by ourselves in Python. 

The parameter settings for each dataset are briefly listed in Table~\ref{tab:setting}. For a fair comparison, the parameter $ \nu $ in $ \nu $-SVM (with a linear kernel) is fixed for all methods on each dataset. For six competing methods, we choose the optimal hyperparameters according to their original papers. For our GDM model, a linear kernel is fixed, while the influence of different kernels are shown in Figs.~S3-S8 in the \emph{Supplementary File}. For the Raider dataset, we set $ G_{ij} = 1 $ if the $ i $-th and $ j $-th samples are temporally aligned; and $ G_{ij} = 0 $, otherwise. For the other datasets, we set $ G_{ij} = 1 $ if the $ i $-th and $ j $-th samples are in the same category; and $ G_{ij} = -1 $, otherwise.

\paragraph{Results on Aligned Datasets}
On the temporally-aligned datasets, we report the BSC accuracy values achieved by eight different methods in Table~\ref{tab:result}. As can be seen from Table~\ref{tab:result}, for each aligned dataset, the proposed GDM method consistently outperforms the competing methods in terms of BSC accuracy. For example, GDM achieves the improvement of $>8\%$ compared to the second best result (i.e., $52.50$ of RHA) on the DIS105WB dataset.

\paragraph{Results on Unaligned Datasets}
To assess the performance of GDM when dealing with unaligned data, we randomly remove some data from each aligned dataset. Here, the term $ q\% $ incompleteness means that $ q $ percent of aligning data are randomly removed per subject. The corresponding results are shown in Fig.~\ref{fig:incomplete}. Notably, six competing methods (i.e., HA, RHA, KHA, SVDHA, SRM, and RSRM) cannot be applied to such incomplete datasets since they are designed for aligned data. More results can be found in Figs.~S3-S5 in the \emph{Supplementary File}. From Fig.~\ref{fig:incomplete}, one can observe that our GDM is able to preserve a dominant BSC accuracy with incompleteness up to at least $ 20\% $. For the DS232 dataset, the performance of GDM still beats others with $ 50\% $ incompleteness. These results further validate the superiority of GDM in handling unaligned data.

\paragraph{Necessity of Low-Dimension Assumption}
To evaluate the influence of low-dimension assumption on GDM, we perform another group of experiments to study the BSC values achieved by GDM with different energy ratios kept on three datasets, with results reported in Figure~\ref{fig:energy}. This figure suggests that the best results are not achieved by GDM with $ 100\% $ energy on each dataset, thus verifying the importance of the low-dimension assumption. Besides, on the Raider dataset, GDM still achieves a good result with around $ 20\% $ energy kept. We conjecture that it results from the fact that the movie data contain much richer information than the visual data generated from simple objects. More results can be found in Figs.~S6-S8 in the \emph{Supplementary File}.

\section{Conclusion}
As an essential step in fMRI analysis, functional alignment removes the differences between subjects' brains so that multi-subject fMRI data can be aggregated to make valid and general inferences. However, the existing methods cannot well handle unaligned fMRI datasets. 
In this paper, a flexible framework is developed on a cross-subject graph that depicts the (dis)similarities among all samples. To reduce the computational cost, the framework is regularized so that a novel feasible kernel-based optimization is analytically developed. To avoid overfitting caused by the HSLT resolution of fMRI, a low-dimension assumption is made over each new feature space, and we also propose a way to incorporate such an assumption into our proposed optimization.
Experimental results attest to the superiority of GDM. In the future, we plan to study how to construct an informative graph matrix in different situations. 

\section{Acknowledgments} 
This work was in part supported by the National Natural Science Foundation of China (Nos. 61876082, 61732006, 61861130366, 61703301), the National Key R\&D Program of China (Nos. 2018YFC2001600, 2018YFC2001602), the Taishan Scholar Program of Shandong Province in China, and the Shandong Natural Science Foundation for Distinguished Young Scholar in China (No.~ZR2019YQ27). 

\fontsize{9.6pt}{10.6pt} \selectfont

\bibliography{aaai2020}
\bibliographystyle{aaai}



\section*{Supplementary material}

# Graph-Based Decoding Model for Functional Alignment of Unaligned fMRI Data
## – *Supplementary File*

Weida Li, Mingxia Liu, Fang Chen, Daoqiang Zhang

In what follows, we will first introduce experimental setup used in the main text, and then present additional results achieved by the proposed Graph-based Decoding Model (GDM) using different kernels. Then, we present rigorous and comprehensive proofs for GDM and some basic proofs needed for proving GDM in an appendix section.

# 1 Experimental Scheme

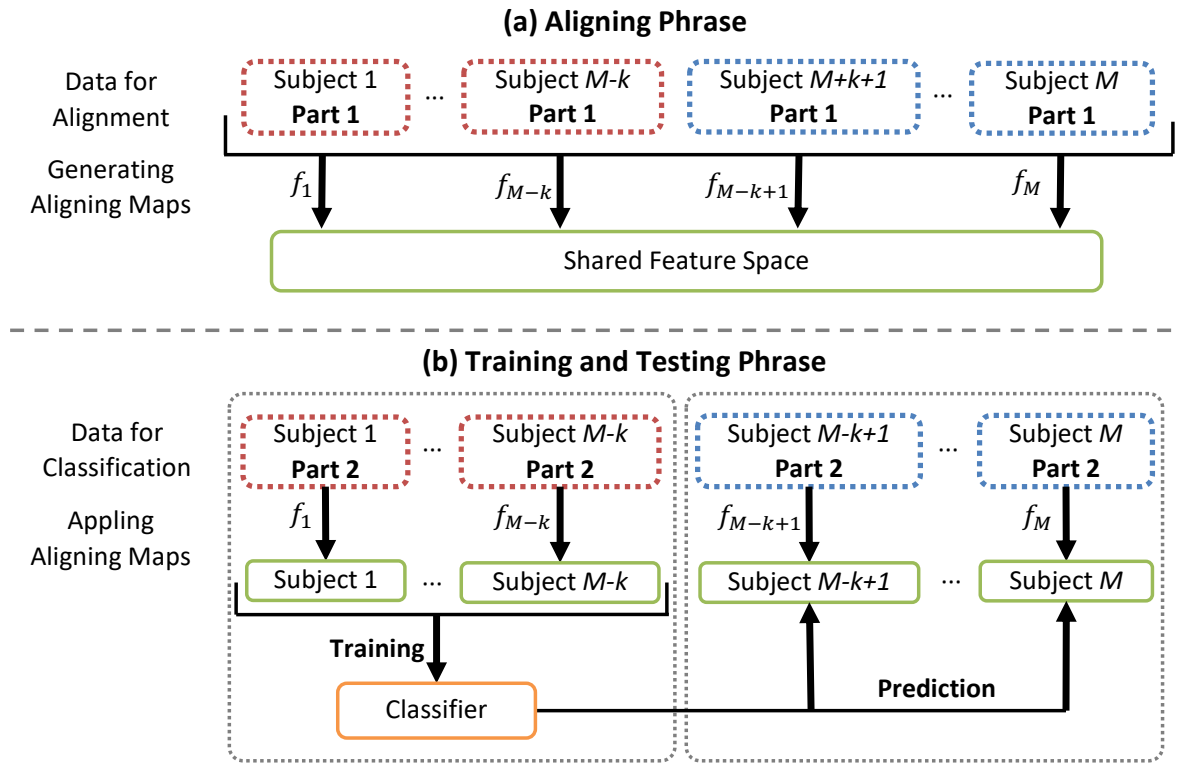

Figure S1: The paradigm of each experiment. Here, leave-k-subject-out strategy is taken. Specifically, the aligning data are never used in the training and testing phrase.

# 2 Influence of Different Kernels

In the main text, we empirically use a linear kernel in GDM. In the following, we study the influence of different kernels on the performance of GDM. These kernels are listed as follows.

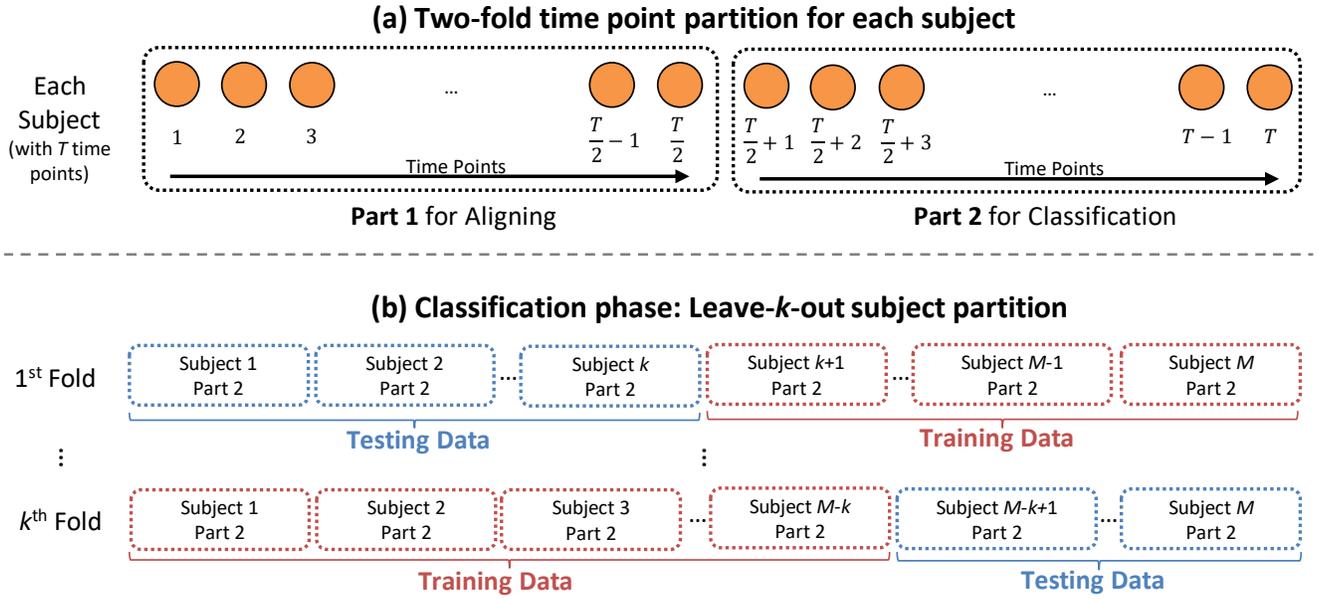

Figure S2: The cross-validation strategy used in our experiments. Here, it demonstrates how we divide subjects with leave-k-subject-out strategy.

$$\text{Gaussian kernel } k(\mathbf{x}, \mathbf{y}) = \exp\left(-\frac{\|\mathbf{x} - \mathbf{y}\|^2}{hp}\right),$$

$$\text{quadratic kernel } k(\mathbf{x}, \mathbf{y}) = \left(\mathbf{x}^T \mathbf{y} + hp\right)^2,$$

$$\text{sigmoid kernel } k(\mathbf{x}, \mathbf{y}) = \tanh\left(\frac{\mathbf{x}^T \mathbf{y}}{hp}\right).$$

where $hp$ serves as hyperparameter for each kernel.

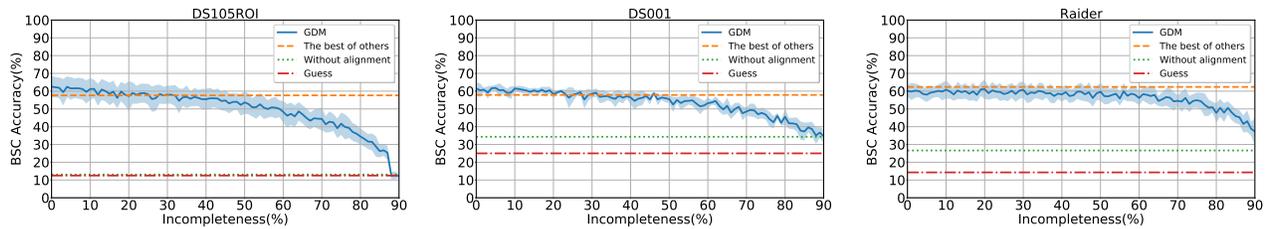

Figure S3: Results of our GDM with a Gaussian kernel ($hp = 5,000, K = 10$) with different incomplete data rates. The energy is $0.65, 0.6, 0.35$ from left to right, $\nu$ is $0.8, 0.5, 0.5$, respectively. "The best of others" refers to the best result of six competing methods with complete data. The term "Without alignment" denotes the $\nu$-SVM method without any alignment. "Guess" denotes the randomly guess method.

## 2.1 Additional Results on Incomplete Dataset

Besides the results in Fig. 1 in the main text, we further report more results on incomplete or unaligned datasets, by using different kernels. The experimental results are shown in Figs. S3-S5. Denote $K$ by the number of the shared features and $\nu$ is for $\nu$-SVM. For each figure, $q\%$ incompleteness means that $q\%$ of the aligning data are randomly removed per subject.

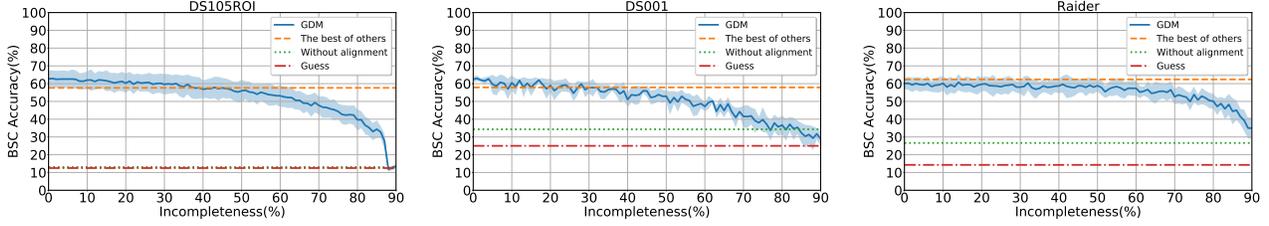

Figure S4: Results of our GDM with a quadratic kernel ($hp = 100, K = 10$) with different incomplete data rates. The energy is $0.8, 0.7, 0.35$, respectively, $\nu$ is $0.8, 0.5, 0.5$, respectively. "The best of others" refers to the best result of six competing methods with complete data. The term "Without alignment" denotes the $\nu$-SVM method without any alignment. "Guess" denotes the randomly guess method.

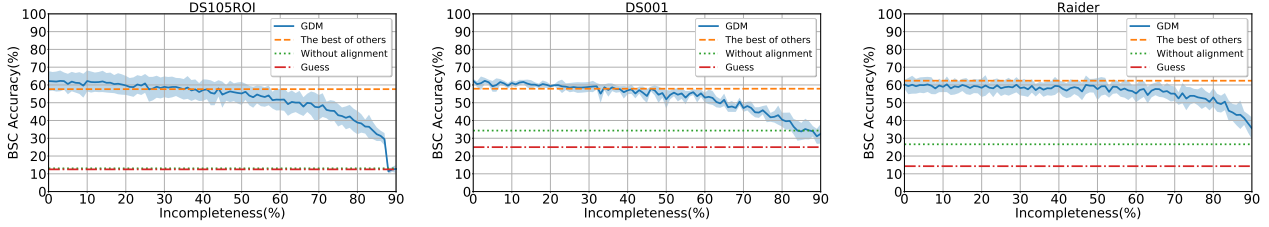

Figure S5: Results of our GDM with a sigmoid kernel ($hp = 200, K = 10$) with different incomplete data rates. The energy is $0.75, 0.7, 0.35$, respectively, $\nu$ is $0.8, 0.5, 0.5$, respectively. "The best of others" refers to the best result of six competing methods with complete data. The term "Without alignment" denotes the $\nu$-SVM method without any alignment. "Guess" denotes the randomly guess method.

## 2.2 Additional Results with Low-Dimension Assumption

In the main text, we evaluate the influence of low-dimension assumption on GDM with a linear kernel on three datasets in Fig. 2. Here, we show more results of GDM with the other three types of kernels using the low-dimension assumption in Figs. S6-S8. In each figure, the term $p\%$ energy shows how much energy is nearly kept per subject.

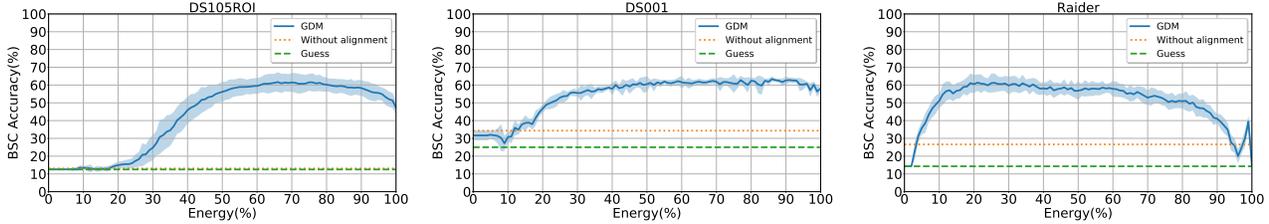

Figure S6: Results of our GDM with a Gaussian kernel ($hp = 5000, K = 6$) with different energy kept. $\nu$ is $0.8, 0.5, 0.5$ from left to right. The term "Without alignment" denotes the $\nu$-SVM method without any alignment. "Guess" denotes the randomly guess method.

# 3 Proofs for GDM

## 3.1 Preliminaries

**Notation and Definitions**   We focus on real Hilbert space $\mathcal{H}$.

1. The bold letters are for matrices (upper) or vectors (lower), while the plain are for scalars.

2. Given a matrix $\mathbf{A}$, $\mathbf{a}_i$ refers to its $i$-th column.

3. Given any sequence of matrices $\{\mathbf{A}^{(i)}\}_{i=1}^M$, let $\mathbf{A}_*$ denotes the corresponding block diagonal matrix whose diagonal matrices are $\{\mathbf{A}^{(i)}\}_{i=1}^M$ from the top left to the bottom right.

3

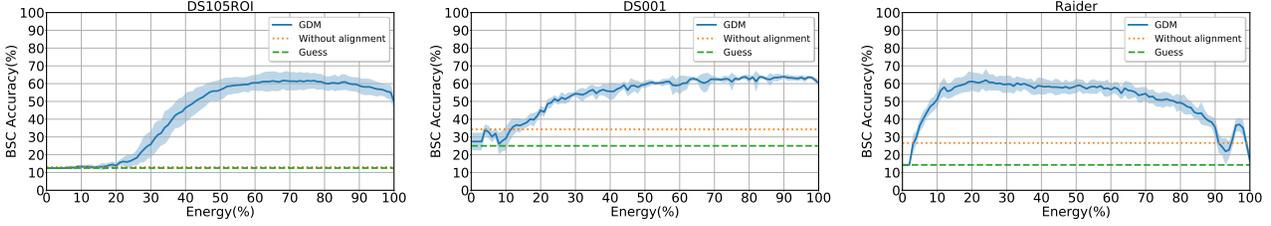

Figure S7: Results of our GDM with a quadratic kernel, $hp = 100, K = 10$) with different energy kept. $\nu$ is $0.8, 0.5, 0.5$, respectively. The term "Without alignment" denotes the $\nu$-SVM method without any alignment. "Guess" denotes the randomly guess method.

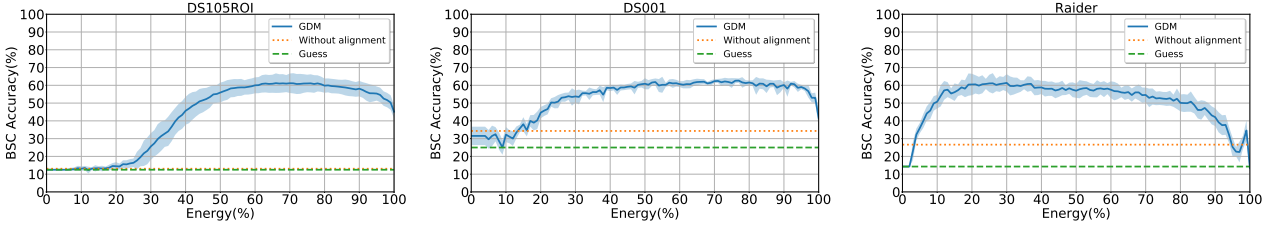

Figure S8: Results of our GDM with a sigmoid kernel ($hp = 200, K = 10$) with different energy kept. $\nu$ is $0.8, 0.5, 0.5$, respectively. The term "Without alignment" denotes the $\nu$-SVM method without any alignment. "Guess" denotes the randomly guess method.

4. Any vector that belongs to a specific finite Euclidean space is treated as a column vector.

5. Bold upper letters with a line above, e.g., $\overline{\mathbf{A}} \in \mathcal{H}^L$, indicates a $L$-tuple of vectors that belong to the real Hilbert space $\mathcal{H}$. Particularly, we also write $\overline{\mathbf{A}}$ as $(\overline{\mathbf{a}}_i)_{1 \le i \le L}$ where $\overline{\mathbf{a}}_i$ refers to the corresponding vector in $\overline{\mathbf{A}}$. Moreover, bold lower letters are for any vector, e.g., $\overline{\mathbf{a}}$, in a Hilbert space.

6. For simplicity, we abuse the notation of transpose and matrix multiplication to denote the inner products between $\overline{\mathbf{A}}$ and $\overline{\mathbf{B}}$, i.e., $\left(\overline{\mathbf{A}}^T \overline{\mathbf{B}}\right)_{ij} = \langle \overline{\mathbf{a}}_i, \overline{\mathbf{b}}_j \rangle$. With this definition, $(\overline{\mathbf{A}}^T \overline{\mathbf{B}})^T = \overline{\mathbf{B}}^T \overline{\mathbf{A}}$.

7. With $\overline{\mathbf{A}}, \overline{\mathbf{B}} \in \mathcal{H}^L$ and $\alpha \in \mathbb{R}$, $\overline{\mathbf{A}} + \overline{\mathbf{B}}$ and $\alpha \overline{\mathbf{A}}$ are defined by element-wise addition and scalar multiplication, respectively. Given a map or function $T$ mapping from $\mathcal{H}$, and $\overline{\mathbf{X}} \in \mathcal{H}^L$, define $T(\overline{\mathbf{X}})$ by $(T(\overline{\mathbf{x}}_i))_{1 \le i \le L}$.

8. Suppose $\overline{\mathbf{A}} \in \mathcal{H}^N, \mathbf{B} \in \mathbb{R}^{N \times L}$ and $\mathbf{C} \in \mathbb{R}^{L \times P}$, we define $\overline{\mathbf{Y}} = \overline{\mathbf{A}} \mathbf{B} \in \mathcal{H}^L$ by setting $\overline{\mathbf{y}}_i = \sum_{j=1}^N B_{ji} \overline{\mathbf{a}}_j$. With such definition, one can check that $(\overline{\mathbf{A}} \mathbf{B}) \mathbf{C} = \overline{\mathbf{A}} (\mathbf{B} \mathbf{C})$. Therefore, we write $\overline{\mathbf{A}} \mathbf{B} \mathbf{C}$ without ambiguity.

9. Given $\overline{\mathbf{X}} \in \mathcal{H}^L$, let $\text{span}(\overline{\mathbf{X}})$ be $\{\sum_{i=1}^L \alpha_i \overline{\mathbf{x}}_i : \boldsymbol{\alpha} \in \mathbb{R}^L\}$, and $\text{span}(\overline{\mathbf{X}})^\perp$ be its orthogonal complement.

**Reserved Letters** Let $\{\mathbf{X}^{(i)} \in \mathbb{R}^{V_i \times T_i}\}_{i=1}^M$ be a fMRI dataset where $T_i$ and $V_i$ are the number of samples (time points or volumes) and features (voxels) of the $i$-th subject, respectively, and $M$ is the total number of subjects. Due to the high-spacial-low-temporal resolution of fMRI, $T_i \ll V_i$. To develop kernel-based method, we introduce a column-wise non-linear map $\Phi_i$ that maps each sample, e.g., each column of $\mathbf{X}^{(i)}$, of the $i$-th subject into a new feature space $\mathcal{H}_i$. For simplicity, let $\overline{\boldsymbol{\Phi}}^{(i)}$ be $\Phi_i(\mathbf{X}^{(i)}) \in \mathcal{H}_i^{T_i}$, i.e., $\overline{\boldsymbol{\phi}}_j^{(i)} = \Phi_i(\mathbf{x}_j^{(i)})$, and $\mathbf{K}^{(i)}$ be $(\overline{\boldsymbol{\Phi}}^{(i)})^T \overline{\boldsymbol{\Phi}}^{(i)}$. Further, let $K$ denote the number of shared features across subjects.

**Problem Statements** The goal is to learn aligning maps $\{f_i : \mathbb{R}^{V_i} \mapsto \mathbb{R}^K\}_{i=1}^M$ for each subject such that they map a population of subjects' fMRI responses into a shared space in which the disparities between subjects' brains are eliminated. Here, we aim to learn linear maps $\{h_i : \mathcal{H}_i \mapsto \mathbb{R}^K\}$ with good generalization. Therefore, $f_i = h_i \circ \Phi_i$ and $h_i(\overline{\boldsymbol{\phi}}_j^{(i)}) = (\overline{\mathbf{W}}^{(i)})^T \overline{\boldsymbol{\phi}}_j^{(i)}$ for $1 \le j \le T_i$ where $\overline{\mathbf{W}}^{(i)} \in \mathcal{H}_i^K$.

To use the graph matrix $\mathbf{G}$ defined in the main paper, we need to cast $\{\overline{\boldsymbol{\Phi}}^{(i)}\}_{i=1}^M$ into a common space $\mathcal{H}_{com} = \mathcal{H}_1 \times \mathcal{H}_2 \times \cdots \times \mathcal{H}_M$ (Cartesian product), which is a Hilbert space with an inner product defined by setting $\langle (\overline{\mathbf{x}}_i)_{1 \le i \le M}, (\overline{\mathbf{y}}_i)_{1 \le i \le M} \rangle =$

$\sum_{i=1}^{M} \langle \overline{\mathbf{x}}_i, \overline{\mathbf{y}}_i \rangle_i$ where $\overline{\mathbf{x}}_i, \overline{\mathbf{y}}_i \in \mathcal{H}_i$ and $\langle \cdot, \cdot \rangle_i$ is the coupled inner product of $\mathcal{H}_i$. For each $i$, define a map $C_i : \mathcal{H}_i \mapsto \mathcal{H}_{com}$ by setting $C_i(\overline{\mathbf{x}}) = (\overline{\mathbf{y}}_j)_{1 \leq j \leq M}$ such that $\overline{\mathbf{y}}_j = \overline{\mathbf{x}}$ if $j = i$ and $\mathbf{0} \in \mathcal{H}_j$ otherwise. Then, let $\overline{\boldsymbol{\Psi}}^{(i)}$ be $C_i\left(\overline{\boldsymbol{\Phi}}^{(i)}\right)$, after which, we group them together by letting $\overline{\boldsymbol{\Psi}}$ be $(\overline{\boldsymbol{\psi}}_1^{(1)}, \ldots, \overline{\boldsymbol{\psi}}_{T_1}^{(1)}, \overline{\boldsymbol{\psi}}_1^{(2)}, \ldots, \overline{\boldsymbol{\psi}}_{T_2}^{(2)}, \ldots, \overline{\boldsymbol{\psi}}_1^{(M)}, \ldots, \overline{\boldsymbol{\psi}}_{T_M}^{(M)})$. Thus, $\overline{\boldsymbol{\Psi}} \in \mathcal{H}_{com}^T$ where $T = \sum_{i=1}^{M} T_i$. Moreover, define $\overline{\mathbf{W}} \in \mathcal{H}_{com}^K$ by setting $\overline{\mathbf{w}}_i = (\overline{\mathbf{w}}_i^{(1)}, \overline{\mathbf{w}}_i^{(2)}, \ldots, \overline{\mathbf{w}}_i^{(M)})$.

### 3.2 The Development of GDM

Since $\mathbf{Y} = \overline{\mathbf{W}}^T \overline{\boldsymbol{\Psi}} = \left( (\overline{\mathbf{W}}^{(1)})^T \overline{\boldsymbol{\Phi}}^{(1)} \quad (\overline{\mathbf{W}}^{(2)})^T \overline{\boldsymbol{\Phi}}^{(2)} \quad \cdots \quad (\overline{\mathbf{W}}^{(M)})^T \overline{\boldsymbol{\Phi}}^{(M)} \right)$ contains all transformed samples across subjects, like the one in the main paper, the crude framework can be expressed as

$$\operatorname*{argmin}_{\overline{\mathbf{W}}} \operatorname{tr}\left(\mathbf{Y}\mathbf{L}\mathbf{Y}^T\right) = \operatorname{tr}\left((\overline{\mathbf{W}}^T \overline{\boldsymbol{\Psi}})\mathbf{L}(\overline{\boldsymbol{\Psi}}^T \overline{\mathbf{W}})\right)$$
$$\text{subject to } (\overline{\mathbf{W}}^T \overline{\boldsymbol{\Psi}})(\overline{\boldsymbol{\Psi}}^T \overline{\mathbf{W}}) = \mathbf{I} \, . \tag{1}$$

**Proposition 1** *If $\overline{\mathbf{W}}$ is one solution for problem (1), then there must be another solution that belongs to* $\operatorname{span}(\overline{\boldsymbol{\Psi}})$ *and has the same objective value as $\overline{\mathbf{W}}$.*

*Proof.* As is proved by Lemma 1, $\operatorname{span}(\overline{\boldsymbol{\Psi}})$ is closed. By Theorem 4.11 in [1], $\mathcal{H}_{com} = \operatorname{span}(\overline{\boldsymbol{\Psi}}) \oplus \operatorname{span}(\overline{\boldsymbol{\Psi}})^\perp$, i.e., for any $\overline{\mathbf{x}} \in \mathcal{H}_{com}$, it can be uniquely decomposed as $r(\overline{\mathbf{x}}) + n(\overline{\mathbf{x}})$ where $r(\overline{\mathbf{x}}) \in \operatorname{span}(\overline{\boldsymbol{\Psi}})$ and $n(\overline{\mathbf{x}}) \in \operatorname{span}(\overline{\boldsymbol{\Psi}})^\perp$. Therefore, $\overline{\mathbf{W}}$ can be unique decomposed as $r(\overline{\mathbf{W}}) + n(\overline{\mathbf{W}})$. With

$$\overline{\mathbf{W}}^T \overline{\boldsymbol{\Psi}} = (r(\overline{\mathbf{W}}) + n(\overline{\mathbf{W}}))^T \overline{\boldsymbol{\Psi}} = r(\overline{\mathbf{W}})^T \overline{\boldsymbol{\Psi}} + n(\overline{\mathbf{W}})^T \overline{\boldsymbol{\Psi}} = r(\overline{\mathbf{W}})^T \overline{\boldsymbol{\Psi}} + \mathbf{0} = r(\overline{\mathbf{W}})^T \overline{\boldsymbol{\Psi}} \, ,$$

$r(\overline{\mathbf{W}}) \in \operatorname{span}(\overline{\boldsymbol{\Psi}})$ satisfies the constraint in problem (1) and it shares the same objective value with $\overline{\mathbf{W}}$. $\square$

Therefore, the framework can be regularized as

$$\operatorname*{argmin}_{\overline{\mathbf{W}}} \operatorname{tr}\left(\mathbf{Y}\mathbf{L}\mathbf{Y}^T\right) = \operatorname{tr}\left((\overline{\mathbf{W}}^T \overline{\boldsymbol{\Psi}})\mathbf{L}(\overline{\boldsymbol{\Psi}}^T \overline{\mathbf{W}})\right)$$
$$\text{subject to } (\overline{\mathbf{W}}^T \overline{\boldsymbol{\Psi}})(\overline{\boldsymbol{\Psi}}^T \overline{\mathbf{W}}) = \mathbf{I}$$
$$\overline{\mathbf{w}}_i \in \operatorname{span}(\overline{\boldsymbol{\Psi}}) \text{ for } 1 \leq i \leq K \, . \tag{2}$$

### 3.3 The Development of Kernel-Based Optimization

For each $i$, by spectral decomposition, $(\overline{\boldsymbol{\Phi}}^{(i)})^T \overline{\boldsymbol{\Phi}}^{(i)} = \mathbf{V}^{(i)} \mathbf{D}^{(i)} (\mathbf{V}^{(i)})^T$ where zero eigenvalues are excluded. With $\overline{\mathbf{U}}^{(i)} = \overline{\boldsymbol{\Phi}}^{(i)} \mathbf{V}^{(i)} (\mathbf{D}^{(i)})^{-\frac{1}{2}}$ and Lemma 2, it leads to an SVD of $\overline{\boldsymbol{\Phi}}^{(i)}$, i.e.,

$$\overline{\boldsymbol{\Phi}}^{(i)} = \overline{\mathbf{U}}^{(i)} (\mathbf{D}^{(i)})^{\frac{1}{2}} (\mathbf{V}^{(i)})^T \, . \tag{3}$$

Like the way we construct $\overline{\boldsymbol{\Psi}}$ by using $\{\overline{\boldsymbol{\Phi}}^{(i)}\}_{i=1}^M$, we set up $\overline{\mathbf{U}}$ by using $\{\overline{\mathbf{U}}^{(i)}\}_{i=1}^M$. Therefore, there is $\overline{\boldsymbol{\Psi}} = \overline{\mathbf{U}} \mathbf{D}_*^{\frac{1}{2}} \mathbf{V}_*^T$, which is an SVD of $\overline{\boldsymbol{\Psi}}$.

Next, we will show that problem (2) is equivalent to

$$\operatorname*{argmin}_{\mathbf{Q}} \operatorname{tr}\left(\mathbf{Q}^T \mathbf{V}_*^T \mathbf{L} \mathbf{V}_* \mathbf{Q}\right)$$
$$\text{subject to } \mathbf{Q}^T \mathbf{Q} = \mathbf{I} \, . \tag{4}$$

Denote the shape of $\mathbf{D}_*$ by $S \times S$. Let $\mathcal{S}$ be $\{\overline{\mathbf{W}} : (\overline{\mathbf{W}}^T \overline{\boldsymbol{\Psi}})(\overline{\boldsymbol{\Psi}}^T \overline{\mathbf{W}}) = \mathbf{I} \text{ and } \overline{\mathbf{w}}_i \in \operatorname{span}(\overline{\boldsymbol{\Psi}}) \text{ for } 1 \leq i \leq K\}$ and $\mathcal{T}$ be $\{\mathbf{Q} \in \mathbb{R}^{S \times K} : \mathbf{Q}^T \mathbf{Q} = \mathbf{I}\}$. Denote a map $g : \mathcal{S} \mapsto \mathcal{T}$ by setting $g(\overline{\mathbf{W}}) = \mathbf{D}_*^{\frac{1}{2}} \left(\overline{\mathbf{U}}^T \overline{\mathbf{W}}\right)$. Since $\overline{\mathbf{w}}_i \in \operatorname{span}(\overline{\boldsymbol{\Psi}}) = \operatorname{span}(\overline{\mathbf{U}})$ for $1 \leq i \leq K$, $\overline{\mathbf{U}} \mathbf{D}_*^{-\frac{1}{2}} \mathbf{D}_*^{\frac{1}{2}} \left(\overline{\mathbf{U}}^T \overline{\mathbf{W}}\right) = \overline{\mathbf{W}}$, which in turn leads to that $g$ is a bijection between $\mathcal{S}$ and $g(\mathcal{S})$. Suffice to show that $g(\mathcal{S}) = \mathcal{T}$.

Suppose $\mathbf{Q} \in \mathcal{T}$, let $\overline{\mathbf{W}}$ be $\overline{\mathbf{U}} \mathbf{D}_*^{-\frac{1}{2}} \mathbf{Q}$, which means $\overline{\mathbf{w}}_i \in \operatorname{span}(\overline{\boldsymbol{\Psi}})$ for $1 \leq i \leq K$. Then $\overline{\mathbf{W}}^T \overline{\boldsymbol{\Psi}} = \left(\overline{\mathbf{U}} \mathbf{D}_*^{-\frac{1}{2}} \mathbf{Q}\right) \left(\overline{\mathbf{U}} \mathbf{D}_*^{\frac{1}{2}} \mathbf{V}^T\right) = \mathbf{Q}^T \mathbf{D}_*^{-\frac{1}{2}} \left(\overline{\mathbf{U}}^T \overline{\mathbf{U}}\right) \mathbf{D}_*^{\frac{1}{2}} \mathbf{V}^T = \mathbf{Q}^T \mathbf{V}^T$. Therefore, $\overline{\mathbf{W}} \in \mathcal{S}$. Moreover, $g(\overline{\mathbf{U}} \mathbf{D}_*^{-\frac{1}{2}} \mathbf{Q}) = \mathbf{D}_*^{\frac{1}{2}} \left(\overline{\mathbf{U}}^T \left(\overline{\mathbf{U}} \mathbf{D}_*^{-\frac{1}{2}} \mathbf{Q}\right)\right) = \mathbf{Q}$. As a result, the equivalence above holds.

**Proposition 2** *By spectral decomposition, $\mathbf{V}_*^T\mathbf{LV}_* = \mathbf{E}\mathbf{\Lambda}\mathbf{E}^T$ where all eigenvalues of $\mathbf{V}_*^T\mathbf{LV}_*$ along the diagonal of $\mathbf{\Lambda}$ from the top left to the bottom right are in ascending order. Denote the shape of $\mathbf{V}_*^T\mathbf{LV}_*$ by $S \times S$. If $K \leq S$, then the first $K$ columns of $\mathbf{E}$ is an optimal solution for problem (4).*

*Proof.* Firstly, problem (4) is equivalent to

$$\operatorname*{argmin} \operatorname{tr}\left(\mathbf{R}^T \mathbf{\Lambda} \mathbf{R}\right) \quad \text{subject to } \mathbf{R}^T\mathbf{R} = \mathbf{I} \quad (5)$$

where $\mathbf{R} = \mathbf{E}^T\mathbf{Q}$. Here, $\mathbf{R}^T\mathbf{R} = \mathbf{I}$ implies $\sum_{i=1}^{S}\sum_{j=1}^{K} R_{ij}^2 = K$ and $\sum_{j=1}^{K} R_{ij}^2 \leq 1$ for each $i$, which in turn leads to

$$\operatorname{tr}\left(\mathbf{R}^T\mathbf{\Lambda}\mathbf{R}\right) = \sum_{i=1}^{S} \Lambda_{ii} \sum_{j=1}^{K} R_{ij}^2 \geq \sum_{i=1}^{K} \Lambda_{ii} \ .$$

With a moment's thought, $\mathbf{R}^* = \begin{pmatrix}\mathbf{I}_{K\times K} & \mathbf{0}_{K\times(S-K)}\end{pmatrix}^T$ is exactly a global solution for problem (5). Therefore, an optimal solution $\mathbf{Q}^* = \mathbf{E}\mathbf{R}^*$ for problem (4) is indeed the first $K$ columns of $\mathbf{E}$. □

Let $\hat{\mathbf{E}}$ denote the first $K$ columns of $\mathbf{E}$, then an optimal solution for problem (2) is

$$\overline{\mathbf{W}}^* = \overline{\mathbf{U}}\mathbf{D}_*^{-\frac{1}{2}}\hat{\mathbf{E}} = \overline{\mathbf{\Psi}}\mathbf{V}_*\mathbf{D}_*^{-1}\hat{\mathbf{E}} \ . \quad (6)$$

Since each $\overline{\mathbf{W}}^{(i)}$ is separable from $\overline{\mathbf{W}}$, an optimal solution for subject $i$ is $(\overline{\mathbf{W}}^{(i)})^* = \overline{\mathbf{\Phi}}^{(i)}\mathbf{V}^{(i)}(\mathbf{D}^{(i)})^{-1}\hat{\mathbf{E}}^{(i)}$ where $\{\hat{\mathbf{E}}^{(i)}\}_{i=1}^{M}$ are block matrices of $\hat{\mathbf{E}}$, which is cut along the first dimension according to the dimensions of block matrices in $\mathbf{D}_*$.

**Proposition 3** *Considering problem (5), the shape of $\mathbf{R}$ is $S \times K$. If $K = S$, or $K < S$ with $\Lambda_{KK} < \Lambda_{(K+1)(K+1)}$, then its optimal solution is unique except being rotated. In other words, if $\mathbf{R}^{(1)}$ and $\mathbf{R}^2$ are two optimal solutions, there exists an orthogonal matrix $\mathbf{P}$ such that $\mathbf{R}^{(1)} = \mathbf{R}^{(2)}\mathbf{P}$. By the equivalence between problem (5) and (2), the optimal solution of problem (2) has such uniqueness.*

*Proof.* Note that the diagonal elements along $\mathbf{\Lambda}$ are in ascending order and

$$\operatorname{tr}\left(\mathbf{R}^T\mathbf{\Lambda}\mathbf{R}\right) = \sum_{i=1}^{S}\Lambda_{ii}\sum_{j=1}^{K} R_{ij}^2 \geq \sum_{i=1}^{K}\Lambda_{ii} \ .$$

If $K = S$, then any orthogonal matrix has the same objective value since $\operatorname{tr}\left(\mathbf{R}^T\mathbf{\Lambda}\mathbf{R}\right) = \sum_{i=1}^{S}\Lambda_{ii}\sum_{j=1}^{S}R_{ij}^2 = \sum_{i=1}^{S}\Lambda_{ii}$.

Suppose $K < S$ and $\Lambda_{KK} < \Lambda_{(K+1)(K+1)}$. For each $\mathbf{R}$, we write it as $\mathbf{R}^T = \begin{pmatrix}\mathbf{S}_{K\times K} & \mathbf{T}_{K\times(S-K)}\end{pmatrix}$. Suffice it to show that $\mathbf{T} = \mathbf{0}$ if and only if $\mathbf{R}$ is optimal. Suppose $T_{ij} \neq 0$, then $\sum_{i=1}^{K}\sum_{j=1}^{K} R_{ij}^2 < K$. So,

$$\sum_{i=1}^{S}\Lambda_{ii}\sum_{j=1}^{K} R_{ij}^2 = \sum_{i=1}^{K}\Lambda_{ii}\sum_{j=1}^{K} R_{ij}^2 + \sum_{i=K+1}^{S}\Lambda_{ii}\sum_{j=1}^{K} R_{ij}^2 > \sum_{i=1}^{K}\Lambda_{ii}$$

as $\Lambda_{KK} < \Lambda_{ii}$ for $K+1 \leq i \leq S$ and $\sum_{i=1}^{S}\sum_{j=1}^{K} R_{ij}^2 = K$. If $\mathbf{T} = \mathbf{0}$, then $\sum_{j=1}^{K} R_{ij}^2 = 1$ for $1 \leq i \leq K$, which means that $\mathbf{R}$ is optimal.

Therefore, any optimal solution $(\mathbf{R}^*)^T$ can be expressed as $\begin{pmatrix}\mathbf{S}_{K\times K} & \mathbf{0}\end{pmatrix}$. $\mathbf{S}$ must be orthogonal since $\mathbf{SS}^T = \mathbf{R}^T\mathbf{R} = \mathbf{I}$.

By the equivalences, any optimal solution $\mathbf{Q}^*$ of problem (4) can be written as $\mathbf{Q}^* = \mathbf{E}\mathbf{R}^* = \hat{\mathbf{E}}\mathbf{S}^T$ where $\hat{\mathbf{E}}$ is the first $K$ columns of $\mathbf{E}$, which in turn shows that any optimal solution $\overline{\mathbf{W}}^*$ can be represented by $\overline{\mathbf{U}}\mathbf{D}_*^{-\frac{1}{2}}\mathbf{Q}^* = \overline{\mathbf{\Psi}}\mathbf{V}_*\mathbf{D}_*^{-1}\hat{\mathbf{E}}\mathbf{S}^T$. Therefore, $(\overline{\mathbf{W}}^{(i)})^* = \overline{\mathbf{\Phi}}^{(i)}\mathbf{V}^{(i)}(\mathbf{D}^{(i)})^{-1}\hat{\mathbf{E}}^{(i)}\mathbf{S}^T$ where $\{\hat{\mathbf{E}}^{(i)}\}_{i=1}^{M}$ are block matrices of $\hat{\mathbf{E}}$, which is cut along the first dimension according to the dimensions of block matrices in $\mathbf{D}_*$. □

### 3.4 The Low-Dimension Assumption

For the $i$-th subject, the related problem is

$$\operatorname*{argmin}_{\overline{\mathbf{m}}_i \in \mathcal{H}_i, \overline{\mathbf{F}}^{(i)} \in \mathcal{H}_i^{L_i}} \sum_{j=1}^{T_i} \left\| \overline{\mathbf{F}}^{(i)}\left((\overline{\mathbf{F}}^{(i)})^T\left(\overline{\phi}_j^{(i)} - \overline{\mathbf{m}}_i\right)\right) - \left(\overline{\phi}_j^{(i)} - \overline{\mathbf{m}}_i\right) \right\|^2 \quad (7)$$

$$\text{subject to } (\overline{\mathbf{F}}^{(i)})^T \overline{\mathbf{F}}^{(i)} = \mathbf{I} \ . \quad (8)$$

By Lemma 4, an optimal solution is $\mathbf{m}_i^* = T_i^{-1} \sum_{j=1}^{T_i} \boldsymbol{\phi}_j^{(i)}$ and $\mathbf{F}_i^*$ be the first $L_i$ columns of $\overline{\mathbf{U}}^{(i)}$ in Eq. (3) where $\boldsymbol{\phi}_j^{(i)} \leftarrow \boldsymbol{\phi}_j^{(i)} - \mathbf{m}_i^*$ for $1 \leq j \leq T_i$.

From now on, suppose all Gram matrices have been centralized. As is provided in Eq. (3), $\overline{\boldsymbol{\Phi}}^{(i)} = \overline{\mathbf{U}}^{(i)} (\mathbf{D}^{(i)})^{\frac{1}{2}} (\mathbf{V}^{(i)})^T$, which is an SVD. Denote the number of (non-zero) singular values in $(\mathbf{D}^{(i)})^{\frac{1}{2}}$ by $s_i$. Assume that the singular values in $(\mathbf{D}^{(i)})^{\frac{1}{2}}$ are in descending order and the first $L_i$ ($L_i \leq s_i$) singular values approximately contains $p_i\%$ ($p_i \in (0, 100]$) energy, i.e., $\sum_{j=1}^{L_i} (D_{jj}^{(i)})^{\frac{1}{2}} / \sum_{j=1}^{s_i} (D_{jj}^{(i)})^{\frac{1}{2}} \approx p_i\%$. Let $\hat{\overline{\mathbf{U}}}^{(i)}$ be the first $L_i$ vectors in $\overline{\mathbf{U}}^{(i)}$, which is an optimal solution of problem (7), and $\hat{\mathbf{V}}^{(i)}$ be the first $L_i$ columns of $\mathbf{V}^{(i)}$. Suppose $\mathbf{Z}^{(i)} \in \mathbf{R}^{V_i \times E_i}$ is some fMRI data for subject $i$, and then let $\overline{\mathbf{Z}}^{(i)}$ be $\Phi_i(\mathbf{Z}^{(i)})$. Note that

$$\left( \hat{\overline{\mathbf{U}}}^{(i)} \left( (\hat{\overline{\mathbf{U}}}^{(i)})^T \overline{\mathbf{Z}}^{(i)} \right) \right)^T \left( \hat{\overline{\mathbf{U}}}^{(i)} \left( (\hat{\overline{\mathbf{U}}}^{(i)})^T \overline{\boldsymbol{\Phi}}^{(i)} \right) \right) = \left( (\hat{\overline{\mathbf{U}}}^{(i)})^T \overline{\mathbf{Z}}^{(i)} \right)^T \left( (\hat{\overline{\mathbf{U}}}^{(i)})^T \overline{\boldsymbol{\Phi}}^{(i)} \right)$$

$$= (\overline{\mathbf{Z}}^{(i)})^T \left( \hat{\overline{\mathbf{U}}}^{(i)} \left( (\hat{\overline{\mathbf{U}}}^{(i)})^T \overline{\boldsymbol{\Phi}}^{(i)} \right) \right)$$

**Proposition 4**

$$\left( (\overline{\mathbf{Z}}^{(i)})^T \overline{\boldsymbol{\Phi}}^{(i)} \right) \hat{\mathbf{V}}^{(i)} = (\overline{\mathbf{Z}}^{(i)})^T \left( \overline{\boldsymbol{\Phi}}^{(i)} \hat{\mathbf{V}}^{(i)} \right) = (\overline{\mathbf{Z}}^{(i)})^T \left( \hat{\overline{\mathbf{U}}}^{(i)} \left( (\hat{\overline{\mathbf{U}}}^{(i)})^T \overline{\boldsymbol{\Phi}}^{(i)} \right) \right) \hat{\mathbf{V}}^{(i)} . \qquad (9)$$

*Proof.* Since

$$\overline{\boldsymbol{\Phi}}^{(i)} \hat{\mathbf{V}}^{(i)} = \overline{\mathbf{U}}^{(i)} (\mathbf{D}^{(i)})^{\frac{1}{2}} (\mathbf{V}^{(i)})^T \hat{\mathbf{V}}^{(i)} = \overline{\mathbf{U}}^{(i)} (\mathbf{D}^{(i)})^{\frac{1}{2}} \begin{pmatrix} \mathbf{I}_{L_i \times L_i} \\ \mathbf{0} \end{pmatrix} = \hat{\overline{\mathbf{U}}}^{(i)} \boldsymbol{\Lambda}^{(i)}$$

where $\boldsymbol{\Lambda}^{(i)}$ is the upper left $L_i \times L_i$ submatrix of $(\mathbf{D}^{(i)})^{\frac{1}{2}}$, there is

$$\left( \hat{\overline{\mathbf{U}}}^{(i)} \left( (\hat{\overline{\mathbf{U}}}^{(i)})^T \overline{\boldsymbol{\Phi}}^{(i)} \right) \right) \hat{\mathbf{V}}^{(i)} = \hat{\overline{\mathbf{U}}}^{(i)} \left( (\hat{\overline{\mathbf{U}}}^{(i)})^T \left( \overline{\boldsymbol{\Phi}}^{(i)} \hat{\mathbf{V}}^{(i)} \right) \right) = \hat{\overline{\mathbf{U}}}^{(i)} \boldsymbol{\Lambda}^{(i)} = \overline{\boldsymbol{\Phi}}^{(i)} \hat{\mathbf{V}}^{(i)} .$$

$\square$

## 4 Appendix

**Lemma 1** *Suppose $\overline{\mathbf{X}} \in \mathcal{H}^N$, then the subspace $\mathrm{span}(\overline{\mathbf{X}}) = \{\sum_{i=1}^N \alpha_i \overline{\mathbf{x}}_i : \alpha_i \in \mathbb{R}\}$ is closed.*

*Proof.* Without loss of generality, suppose $\overline{\mathbf{X}}$ is linearly independent, i.e., $\sum_{i=1}^N \alpha_i \overline{\mathbf{x}}_i = \mathbf{0} \implies \boldsymbol{\alpha}_i = \mathbf{0}$. Let $\mathbf{A}$ be $\overline{\mathbf{X}}^T \overline{\mathbf{X}} \in \mathbb{R}^{N \times N}$. Since $\boldsymbol{\alpha}^T \mathbf{A} \boldsymbol{\alpha} = \langle \sum_{i=1}^N \alpha_i \overline{\mathbf{x}}_i, \sum_{i=1}^N \alpha_i \overline{\mathbf{x}}_i \rangle \geq 0$ for any $\boldsymbol{\alpha} \in \mathbb{R}^N$ and $\mathbf{A}$ is symmetric, $\mathbf{A}$ is positive semi-definite. If $\mathbf{A}\boldsymbol{\alpha} = \mathbf{0}$, then $\boldsymbol{\alpha}^T \mathbf{A} \boldsymbol{\alpha} = 0 \implies \sum_{i=1}^N \alpha_i \overline{\mathbf{x}}_i = \mathbf{0}$. The linear independence of $\overline{\mathbf{X}}$ indicates $\boldsymbol{\alpha} = \mathbf{0}$. So, $\mathbf{A}$ is invertible. By Cholesky Decomposition, $\mathbf{A} = \mathbf{B}^T \mathbf{B}$. The non-singularity of $\mathbf{A}$ ensures the columns of $\mathbf{B}$ form a basis of $\mathbb{R}^N$.

Define a linear map $T : \mathrm{span}(\overline{\mathbf{X}}) \mapsto \mathbb{R}^N$ by setting

$$T(\sum_{i=1}^N \alpha_i \overline{\mathbf{x}}_i) = \sum_{i=1}^N \alpha_i \mathbf{b}_i .$$

One can check that $T$ is a bijection. Specifically, $T$ is an isomorphism since $\langle T(\overline{\mathbf{y}}), T(\overline{\mathbf{z}}) \rangle = \langle \overline{\mathbf{y}}, \overline{\mathbf{z}} \rangle$ for any $\overline{\mathbf{y}}, \overline{\mathbf{z}} \in \mathrm{span}(\overline{\mathbf{X}})$, which implies that $T^{-1}$ is continuous and $T$ is an isometry.

Suppose $\{\overline{\mathbf{x}}_i\}_{i=1}^\infty$ is a sequence in $\mathrm{span}(\overline{\mathbf{X}})$ such that it converges to a point $\overline{\mathbf{x}} \in \mathcal{H}$. Suffice to prove that $\overline{\mathbf{x}} \in \mathrm{span}(\overline{\mathbf{X}})$. Since $\{\overline{\mathbf{x}}_i\}_{i=1}^\infty$ is a Cauchy sequence, so is $\{T(\overline{\mathbf{x}}_i)\}_{i=1}^\infty$. Therefore, there exists a point $\mathbf{c} \in \mathbb{R}^N$ such that $T(\overline{\mathbf{x}}_i) \to \mathbf{c}$ as $i \to \infty$. The continuity of $T^{-1}$ shows that $\overline{\mathbf{x}}_i$ converges to $T^{-1}(\mathbf{c}) \in \mathrm{span}(\overline{\mathbf{X}})$. Since there could not be two limits in any metric space, $T^{-1}(\mathbf{c}) = \overline{\mathbf{x}}$. $\square$

**Lemma 2** *Suppose $\overline{\mathbf{X}} \in \mathcal{H}^N$, then it can be decomposed as $\overline{\mathbf{X}} = \overline{\mathbf{U}} \boldsymbol{\Sigma} \mathbf{V}^T$ where $\mathbf{V}^T \mathbf{V} = \overline{\mathbf{U}}^T \overline{\mathbf{U}} = \mathbf{I}$ and $\boldsymbol{\Sigma}$ is a diagonal matrix in which the diagonal elements are all positive (and can be ordered). we refers to such decomposition as a Singular Value Decomposition (SVD) of $\overline{\mathbf{X}}$.*

*Proof.* Firstly, remove some vectors if need be in $\overline{\mathbf{X}}$ to gain a list $\overline{\mathbf{Y}} \in \mathcal{H}^M$ ($M \leq N$) such that $\overline{\mathbf{Y}}$ is linearly independent and $\text{span}(\overline{\mathbf{X}}) = \text{span}(\overline{\mathbf{Y}})$. By Lemma 1, there is a bijection $T$ between $\text{span}(\overline{\mathbf{Y}})$ and $\mathbb{R}^M$.

By spectral decomposition, $\overline{\mathbf{X}}^T\overline{\mathbf{X}} = T(\overline{\mathbf{X}})^T T(\overline{\mathbf{X}}) = \mathbf{V}\mathbf{D}\mathbf{V}^T$ where the zero eigenvalues are excluded. We can order the diagonal elements along $\mathbf{D}$ if necessary. With $\mathbf{U} = T(\overline{\mathbf{X}})\mathbf{V}\mathbf{D}^{-\frac{1}{2}}$, it leads to an SVD decomposition of $T(\overline{\mathbf{X}})$, i.e., $T(\overline{\mathbf{X}}) = \mathbf{U}\mathbf{D}^{\frac{1}{2}}\mathbf{V}^\mathbf{T}$. Therefore, there is

$$\overline{\mathbf{X}} = T^{-1}(\mathbf{U}\mathbf{D}^{\frac{1}{2}}\mathbf{V}^T) = T^{-1}(\mathbf{U})\mathbf{D}^{\frac{1}{2}}\mathbf{V}^T ,$$

resulting from the fact that $T^{-1}$ is a linear map. Let $\overline{\mathbf{U}}$ be $T^{-1}(\mathbf{U})$ and $\boldsymbol{\Sigma}$ be $\mathbf{D}^{\frac{1}{2}}$. The isomorphism $T$ guarantees that $\overline{\mathbf{U}}^T\overline{\mathbf{U}} = \mathbf{I}$. Particularly,

$$\overline{\mathbf{U}} = T^{-1}(\mathbf{U}) = T^{-1}\left(T(\overline{\mathbf{X}})\mathbf{V}\mathbf{D}^{-\frac{1}{2}}\right) = \overline{\mathbf{X}}\mathbf{V}\mathbf{D}^{-\frac{1}{2}} .$$

□

**Lemma 3** *Suppose $\overline{\mathbf{X}} \in \mathcal{H}^N$, the unique optimal solution to the problem*

$$\underset{\overline{\mathbf{m}} \in \mathcal{H}}{\text{argmin}} \sum_{i=1}^{N} \|\overline{\mathbf{x}}_i - \overline{\mathbf{m}}\|^2$$

*is $\overline{\mathbf{m}}^* = \frac{1}{N}\sum_{i=1}^{N} \overline{\mathbf{x}}_i$.*

*Proof.* Firstly, remove some vectors if need be in $\overline{\mathbf{X}}$ to gain a list $\overline{\mathbf{Y}} \in \mathcal{H}^M$ ($M \leq N$) such that $\overline{\mathbf{Y}}$ is linearly independent and $\text{span}(\overline{\mathbf{X}}) = \text{span}(\overline{\mathbf{Y}})$. By Lemma 1, there is a bijection $T$ between $\text{span}(\overline{\mathbf{X}})$ and $\mathbb{R}^M$. By theorem 4.11 in [1], there exist two linear maps $r$ and $n$ from $\mathcal{H}$ into $\mathcal{H}$ such that $r(\overline{\mathbf{x}}) \in \text{span}(\overline{\mathbf{X}})$ and $n(\overline{\mathbf{x}}) \in \text{span}(\overline{\mathbf{X}})^\perp$ for any $\overline{\mathbf{x}} \in \mathcal{H}$, as $\|\overline{\mathbf{x}}\|^2 = \|r(\overline{\mathbf{x}})\|^2 + \|n(\overline{\mathbf{x}})\|^2$ and $\overline{\mathbf{x}} = r(\overline{\mathbf{x}}) + n(\overline{\mathbf{x}})$. Since

$$\|\overline{\mathbf{x}}_i - \overline{\mathbf{m}}\|^2 = \|\overline{\mathbf{x}}_i - r(\overline{\mathbf{m}})\|^2 + \|n(\overline{\mathbf{m}})\|^2 \geq \|\overline{\mathbf{x}}_i - r(\overline{\mathbf{m}})\|^2 ,$$

an optimal solution must belong to $\text{span}(\overline{\mathbf{X}})$. With the isomorphism $T$ between $\text{span}(\overline{\mathbf{X}})$ and $\mathbb{R}^M$, there is such equivalence

$$\underset{\overline{\mathbf{m}} \in \text{span}(\overline{\mathbf{X}})}{\text{argmin}} \sum_{i=1}^{N} \|\overline{\mathbf{x}}_i - \overline{\mathbf{m}}\|^2 \iff \underset{\mathbf{m} \in \mathbb{R}^M}{\text{argmin}} \sum_{i=1}^{N} \|\mathbf{x}_i - \mathbf{m}\|^2$$

where $T(\overline{\mathbf{x}}_i) = \mathbf{x}_i$ and $T(\overline{\mathbf{m}}) = \mathbf{m}$.

Considering the right-side problem, since it is convex and differentiable, an optimal solution $\mathbf{m}^*$ must satisfies

$$\frac{\partial}{\partial \mathbf{m}} \sum_{i=1}^{N} \|\mathbf{x}_i - \mathbf{m}\|^2 \bigg|_{\mathbf{m}=\mathbf{m}^*} = -2\sum_{i=1}^{N}(\mathbf{x}_i - \mathbf{m}^*) = 0 ,$$

which leads to $\mathbf{m}^* = \frac{1}{N}\sum_{i=1}^{N} \mathbf{x}_i$. Therefore, $\overline{\mathbf{m}}^* = T^{-1}(\mathbf{m}^*) = \frac{1}{N}\sum_{i=1}^{N} \overline{\mathbf{x}}_i$ as $T^{-1}$ is linear. □

**Lemma 4** *Suppose $\overline{\mathbf{X}} \in \mathcal{H}^N$, let $\overline{\mathbf{Y}}$ be $(\overline{\mathbf{x}}_1 - \overline{\mathbf{a}}, \ldots, \overline{\mathbf{x}}_N - \overline{\mathbf{a}})$ where $\overline{\mathbf{a}} = \frac{1}{N}\sum_{i=1}^{N} \overline{\mathbf{x}}_i$. By Lemma 2, $\overline{\mathbf{Y}}$ can be decomposed as $\overline{\mathbf{U}}\boldsymbol{\Sigma}\mathbf{V}^T$, i.e., Singular Value Decomposition, where the diagonal elements along $\boldsymbol{\Sigma}$ are in descending order and be all positive. Denote the shape of $\boldsymbol{\Sigma}$ by $M \times M$. Let $\overline{\mathbf{R}} \in \mathcal{H}^K$ are a variable such that $K \leq M$. Considering the following optimization problem*

$$\underset{\overline{\mathbf{m}} \in \mathcal{H}, \overline{\mathbf{R}} \in \mathcal{H}^K}{\text{argmin}} \sum_{i=1}^{N} \left\|\overline{\mathbf{R}}(\overline{\mathbf{R}}^T (\overline{\mathbf{x}}_i - \overline{\mathbf{m}})) - (\overline{\mathbf{x}}_i - \overline{\mathbf{m}})\right\|^2$$

*subject to $\overline{\mathbf{R}}^T\overline{\mathbf{R}} = \mathbf{I}$ ,*

*let $\overline{\mathbf{R}}^*$ and $\overline{\mathbf{m}}^*$ be the first $K$ vectors in $\overline{\mathbf{U}}$ and $\frac{1}{N}\sum_{i=1}^{N} \overline{\mathbf{x}}_i$, respectively, then $(\overline{\mathbf{R}}^*, \overline{\mathbf{m}}^*)$ is an optimal solution for the problem above.*

*Proof.* The proof is composed of two parts: 1) Firstly, we show that $\overline{\mathbf{m}}^*$ is always an optimal solution whatever $\overline{\mathbf{R}}$ is fixed. 2) Then, we prove that $\overline{\mathbf{R}}^*$ is an optimal solution when $\overline{\mathbf{m}}$ is fixed as $\overline{\mathbf{m}}^*$.

1) $\overline{\mathbf{m}}^*$ is always optimal independent of $\overline{\mathbf{R}}$.

By Lemma 1, $\mathrm{span}(\overline{\mathbf{R}})$ is closed. By theorem 4.11 in [1], there exist two linear maps $r$ and $n$ from $\mathcal{H}$ into $\mathcal{H}$ such that $r(\overline{\mathbf{x}}) \in \mathrm{span}(\overline{\mathbf{R}})$ and $n(\overline{\mathbf{x}}) \in \mathrm{span}(\overline{\mathbf{R}})^\perp$ for any $\overline{\mathbf{x}} \in \mathcal{H}$, as $\|\overline{\mathbf{x}}\|^2 = \|r(\overline{\mathbf{x}})\|^2 + \|n(\overline{\mathbf{x}})\|^2$ and $\overline{\mathbf{x}} = r(\overline{\mathbf{x}}) + n(\overline{\mathbf{x}})$. Therefore,

$$\sum_{i=1}^{N} \left\| \overline{\mathbf{R}}(\overline{\mathbf{R}}^T (\overline{\mathbf{x}}_i - \overline{\mathbf{m}})) - (\overline{\mathbf{x}}_i - \overline{\mathbf{m}}) \right\|^2 = \sum_{i=1}^{N} \left\| \left( \overline{\mathbf{R}} \left( \overline{\mathbf{R}}^T \overline{\mathbf{x}}_i \right) - \overline{\mathbf{x}}_i \right) - \left( \overline{\mathbf{R}} \left( \overline{\mathbf{R}}^T \overline{\mathbf{m}} \right) - \overline{\mathbf{m}} \right) \right\|^2$$

$$= \sum_{i=1}^{N} \|(r(\overline{\mathbf{x}}_i) - \overline{\mathbf{x}}_i) - (r(\overline{\mathbf{m}}) - \overline{\mathbf{m}})\|^2 = \sum_{i=1}^{N} \|n(\overline{\mathbf{x}}_i) - n(\overline{\mathbf{m}})\|^2$$

By Lemma 3, $\sum_{i=1}^{N} \|n(\overline{\mathbf{x}}_i) - n(\overline{\mathbf{m}})\|^2 \geq \sum_{i=1}^{N} \left\| n(\overline{\mathbf{x}}_i) - \frac{1}{N} \sum_{j=1}^{N} n(\overline{\mathbf{x}}_i) \right\|^2$ for any $\overline{\mathbf{m}}$. Then $\overline{\mathbf{m}}^*$ must be an optimal solution since $n(\overline{\mathbf{m}}^*) = \frac{1}{N} \sum_{j=1}^{N} n(\overline{\mathbf{x}}_i)$ whatever $n$ is.

2) The optimal solution of $\overline{\mathbf{R}}$ when $\overline{\mathbf{m}}$ is fixed as $\overline{\mathbf{m}}^*$. There is

$$\left\| \overline{\mathbf{R}} \left( \overline{\mathbf{R}}^T \overline{\mathbf{y}}_i \right) - \overline{\mathbf{y}}_i \right\|^2 = \langle \overline{\mathbf{R}} \left( \overline{\mathbf{R}}^T \overline{\mathbf{y}}_i \right), \overline{\mathbf{R}} \left( \overline{\mathbf{R}}^T \overline{\mathbf{y}}_i \right) \rangle + \|\overline{\mathbf{y}}_i\|^2 - 2\langle \overline{\mathbf{R}} \left( \overline{\mathbf{R}}^T \overline{\mathbf{y}}_i \right), \overline{\mathbf{y}}_i \rangle.$$

Since $\overline{\mathbf{R}}^T \overline{\mathbf{R}} = \mathbf{I}$, $\langle \overline{\mathbf{R}} \left( \overline{\mathbf{R}}^T \overline{\mathbf{y}}_i \right), \overline{\mathbf{R}} \left( \overline{\mathbf{R}}^T \overline{\mathbf{y}}_i \right) \rangle = \langle \overline{\mathbf{R}}^T \overline{\mathbf{y}}_i, \overline{\mathbf{R}}^T \overline{\mathbf{y}}_i \rangle$. By Theorem 4.22 in [1], $\overline{\mathbf{R}}$ can be extended into a basis of $\mathcal{H}$, by which it is obvious that $\langle \overline{\mathbf{R}} \left( \overline{\mathbf{R}}^T \overline{\mathbf{y}}_i \right), \overline{\mathbf{y}}_i \rangle = \langle \overline{\mathbf{R}}^T \overline{\mathbf{y}}_i, \overline{\mathbf{R}}^T \overline{\mathbf{y}}_i \rangle$. Therefore, the problem we are solving is equivalent to

$$\underset{\overline{\mathbf{R}} \in \mathcal{H}^K}{\mathrm{argmax}} \sum_{i=1}^{N} \langle \overline{\mathbf{R}}^T \overline{\mathbf{y}}_i, \overline{\mathbf{R}}^T \overline{\mathbf{y}}_i \rangle = \mathrm{tr}\left( \left( \overline{\mathbf{R}}^T \overline{\mathbf{Y}} \right) \left( \overline{\mathbf{Y}}^T \overline{\mathbf{R}} \right) \right)$$

subject to $\overline{\mathbf{R}}^T \overline{\mathbf{R}} = \mathbf{I}$.

By Lemma 2, $\overline{\mathbf{Y}} = \overline{\mathbf{U}} \boldsymbol{\Sigma} \mathbf{V}^T$, i.e., Singular Value Decomposition, where the singular values along $\boldsymbol{\Sigma}$ are in descending order and are all positive. Thus, $\overline{\mathbf{R}}^T \overline{\mathbf{Y}} = \overline{\mathbf{R}}^T \left( \overline{\mathbf{U}} \boldsymbol{\Sigma} \mathbf{V}^T \right) = \left( \overline{\mathbf{R}}^T \overline{\mathbf{U}} \right) \boldsymbol{\Sigma} \mathbf{V}^T$, which leads to

$$L(\overline{\mathbf{R}}) = \mathrm{tr}\left( \left( \overline{\mathbf{R}}^T \overline{\mathbf{Y}} \right) \left( \overline{\mathbf{Y}}^T \overline{\mathbf{R}} \right) \right) = \mathrm{tr}\left( \left( \overline{\mathbf{R}}^T \overline{\mathbf{U}} \right) \boldsymbol{\Sigma}^2 \left( \overline{\mathbf{U}}^T \overline{\mathbf{R}} \right) \right).$$

Let $\mathbf{A}$ be $\overline{\mathbf{R}}^T \overline{\mathbf{U}} \in \mathbb{R}^{K \times M}$. Both $\overline{\mathbf{R}}^T \overline{\mathbf{R}} = \mathbf{I}_{K \times K}$ and $\overline{\mathbf{U}}^T \overline{\mathbf{U}} = \mathbf{I}_{M \times M}$ provide that

$$\sum_{i=1}^{M} A_{ji}^2 \leq 1 \text{ for any } j \text{ and } \sum_{j=1}^{K} \sum_{i=1}^{M} A_{ji}^2 \leq K.$$

With these inequalities, there is

$$\mathrm{tr}\left( \mathbf{A} \boldsymbol{\Sigma}^2 \mathbf{A}^T \right) = \sum_{i=1}^{M} \Sigma_{ii}^2 \sum_{j=1}^{K} A_{ji}^2 \leq \sum_{i=1}^{K} \Sigma_{ii}^2.$$

With a moment's thought, $L(\overline{\mathbf{R}}^*) = \sum_{i=1}^{K} \Sigma_{ii}^2$, which confirms that $\overline{\mathbf{R}}^*$ is an optimal solution. $\square$



\end{document}